\pdfoutput=1

\documentclass[11pt]{article}
\usepackage[table]{xcolor}
\usepackage[]{ACL2023}

\usepackage{booktabs}
\usepackage{makecell}

\usepackage{times}
\usepackage{latexsym}
\usepackage{amsmath, amssymb}
\usepackage{amsfonts}
\usepackage{multirow}
\usepackage{rotating}
\usepackage{booktabs}
\usepackage{amsmath}

\usepackage{graphicx}
\usepackage{subcaption}
\usepackage{mwe}
\usepackage{hyperref} 
\usepackage{algorithm}
\usepackage[noend]{algpseudocode}

\usepackage[T1]{fontenc}

\usepackage[utf8]{inputenc}

\usepackage{microtype}

\usepackage{inconsolata}

\expandafter\def\expandafter\normalsize\expandafter{%
    \normalsize%
    \setlength\abovedisplayskip{1pt}%
    \setlength\belowdisplayskip{1pt}%
}

\usepackage{color}

\title{Do Not Design, Learn: A Trainable Scoring Function \\for Uncertainty Estimation in Generative LLMs}

\author{
   Duygu Nur Yaldiz\textsuperscript{1}\footnotemark[1] \quad 
   Yavuz Faruk Bakman\textsuperscript{1}\footnotemark[1] \quad  
   Baturalp Buyukates\textsuperscript{2} \\
   \bf Chenyang Tao\textsuperscript{3}\footnotemark[2] \quad
   Anil Ramakrishna\textsuperscript{3}\footnotemark[2] \quad
   Dimitrios Dimitriadis\textsuperscript{3}\footnotemark[2] \\
   \bf Jieyu Zhao\textsuperscript{1} \quad
   Salman Avestimehr\textsuperscript{1}\\
\textsuperscript{1}University of Southern California \quad
  \textsuperscript{2}University of Birmingham \quad
  \textsuperscript{3}Amazon AI \\
  \texttt{\{yaldiz, ybakman\}@usc.edu}}

\begin{document}
\maketitle
\renewcommand*{\thefootnote}{\fnsymbol{footnote}}

 \footnotetext[1]{Equal contribution.}
 \footnotetext[2]{This work does not relate to their position at Amazon.}

\renewcommand*{\thefootnote}{\arabic{footnote}}

\begin{abstract}

Uncertainty estimation (UE) of generative large language models (LLMs) is crucial for evaluating the reliability of generated sequences. A significant subset of UE methods utilize token probabilities to assess uncertainty, aggregating multiple token probabilities into a single UE score using a scoring function. Existing scoring functions for probability-based UE, such as length-normalized scoring and semantic contribution-based weighting, are designed to solve certain aspects of the problem but exhibit limitations, including the inability to handle biased probabilities and complex semantic dependencies between tokens. To address these issues, in this work, we propose Learnable Response Scoring (LARS) function, a novel scoring function that leverages supervised data to capture complex dependencies between tokens and probabilities, thereby producing more reliable and calibrated response scores in computing the uncertainty of LLM generations. Our comprehensive experiments across question-answering and arithmetical reasoning tasks with various datasets demonstrate that LARS significantly outperforms existing scoring functions, achieving improvements of up to 16\% AUROC score.\footnote{Code is available at \url{https://github.com/duygunuryldz/LARS} and \url{https://github.com/Ybakman/TruthTorchLM}}

\end{abstract}

\section{Introduction}

\begin{figure*}[]
\begin{center}
\includegraphics[width=0.95\textwidth]{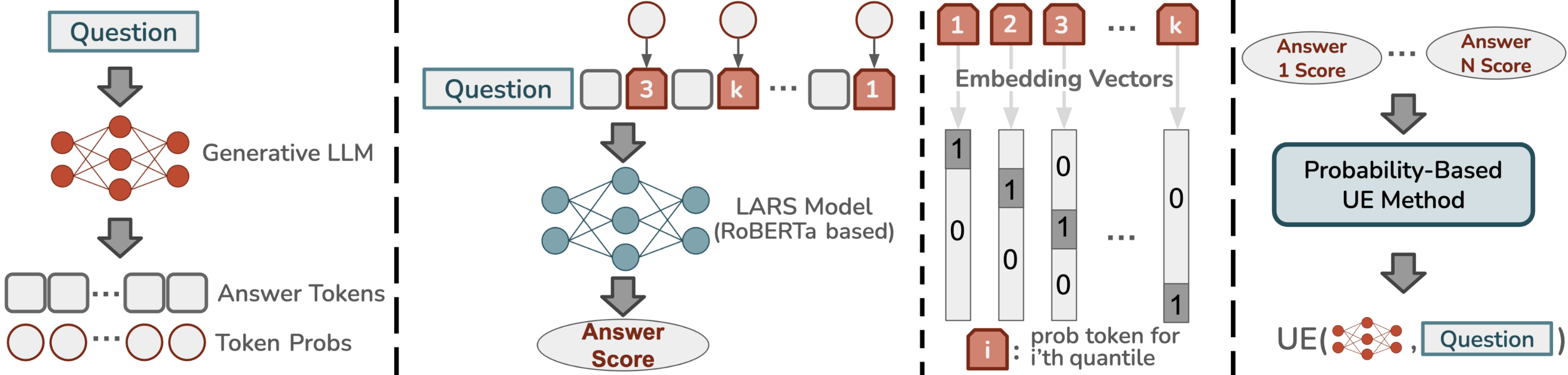}
\vskip -0.1in
\caption{(Left) Answer generation using a generative LLM. (Mid Left) Overview of the proposed scoring function LARS. It utilizes the question, answer tokens, and token probabilities. Token probabilities are fed to LARS model as special probability tokens. (Mid Right) Illustration of few-hot represented embedding vectors of probability tokens. (Right) Overview of probability-based UE methods taking different sampled answer scores such as LNS \cite{malinin2021uncertainty}, MARS \cite{bakman2024mars}, or LARS (this work), and outputting a single UE value.}
\label{fig:lars}
\end{center}
\vskip -0.3in
\end{figure*}

Recent years have seen a transformative shift in AI with the rise of generative Large Language Models (LLMs). Their near-human capabilities in comprehension, generation, and information processing have revolutionized human-machine interactions, driving widespread adoption across industries such as healthcare, law, finance, and marketing \cite{ye2023comprehensive,openai2023gpt4, touvron2023llama2, huang2023benchmarking}. Given that LLMs can sometimes generate misleading or erroneous outputs \cite{ravi2024lynxopensourcehallucination, oguz-etal-2024-llms}, it is crucial to evaluate how much reliance should be placed on their responses. Detecting unreliable, factually incorrect, or irrelevant outputs from LLMs is studied under the topic of hallucination detection \cite{ li-etal-2023-halueval}. Methods such as fact verification \cite{wang2024factcheckbench, chern2023factool}, cross examination \cite{cohen-etal-2023-lm} and Uncertainty Estimation (UE) \cite{malinin2021uncertainty} serve as tools for hallucination detection. 

The field of UE, well-established in classification tasks, has recently been adapted to generative LLMs. In the context of generative LLMs, UE is used to assess the model's reliability for a given query \cite{kuhn2023semantic}. UE methods are particularly valuable as they differ from other hallucination detection approaches by not relying on external resources, such as internet search tools \cite{chern2023factool} or a teacher model \cite{cohen-etal-2023-lm}. UE methods in generative LLMs can be broadly categorized into two categories: 1) Probability-based methods \cite{malinin2021uncertainty, kuhn2023semantic} that utilize token probabilities externally to predict uncertainty. 2) Non-probability-based methods \cite{lyu2024calibrating, chen2024inside} that employ heuristics without relying on token probabilities for estimation. This work focuses on probability-based methods due to their widespread use and promising performance in UE \cite{bakman2024mars, tokensar, kuhn2023semantic}, as well as their applicability to closed-source API models where token probabilities are accessible \cite{openai2023gpt4}.

Probability-based UE in LLMs requires aggregating multiple token probabilities into a single score, which can be done through a scoring function. Length-Normalized Scoring (LNS)~\cite{malinin2021uncertainty, kuhn2023semantic} is a common approach, which calculates the mean of log probabilities of an LLM's output to mitigate bias in longer generations. Subsequent approaches by \citet{bakman2024mars, tokensar} introduce heuristics that prioritize semantically important tokens by assigning higher weights to them, rather than simply averaging as in LNS. However, these scoring functions, largely heuristic in design, often overlook potential pitfalls such as biased probabilities and complex dependencies between tokens. In this work, we critically analyze the weaknesses of the existing scoring functions and introduce a novel scoring function that leverages supervised data to produce more calibrated scores for UE in LLMs.

We summarize our main contributions as follows: \textbf{(1)}~We discuss the limitations of existing scoring functions of UE from three different perspectives including biased probabilities, token dependencies, and applicability to other languages rather than  English. \textbf{(2)}~We introduce a novel off-the-shelf scoring function, Learnable Response Scoring (LARS), which is learned directly from supervised data (visualized in Figure \ref{fig:lars}). \textbf{(3)}~We validate the superiority of LARS over existing baselines across three QA datasets, a mathematical reasoning task, and four different languages. LARS outperforms SOTA scoring functions by up to 16\% in AUROC and 45\% in PRR. Additionally, we analyze its components to explain the effectiveness of LARS.

\section{Preliminaries}\label{background}

\textbf{Uncertainty Estimation in Generative LLMs} addresses the challenge of predicting a model's uncertainty regarding a given input sequence or question. In the context of closed-ended QA and mathematical reasoning tasks, an effective UE method assigns a lower score (indicating less uncertainty) to questions where the model is likely to provide the correct answer (reliable output), and a higher score otherwise. Mathematically, we have $\mathrm{UE}(\theta, x_1) < \mathrm{UE}(\theta, x_2)$ if the most probable generation of model $\theta$ for question $x_1$ is more likely to be correct than for question $x_2$ \cite{malinin2021uncertainty,kuhn2023semantic, tokensar}. 


\textbf{Token Probability-based Methods} use token probabilities to estimate the model uncertainty. This estimation requires aggregating multiple token probabilities into a single score. In their foundational work, \citet{malinin2021uncertainty} formalize the generation's probability for a given question $\mathbf{x}$ and a model parameterized by $\theta$ using the sequence probability. This is defined as follows:
\begin{equation}
    P(\mathbf{s}|\mathbf{x}, \theta) = \prod_{l=1}^{L} P(s_l|s_{<l}, \mathbf{x}; \theta),
\label{normalized-prob}
\end{equation}
where $P(\mathbf{s}|\mathbf{x}, \theta)$ is the  probability for the generated sequence $\mathbf{s}$ (of length $L$), and $s_{<l} \triangleq \{s_1, s_2, \ldots, s_{l-1}\}$ represents the tokens generated before token $s_l$. This sequence probability is used in entropy calculation $\mathcal{H}(\mathbf{x}, \theta)$ by making a Monte Carlo approximation, which requires multiple answer sampling for the given question:
\begin{equation}
\mathcal{H}(\mathbf{x},\theta)\approx - \frac{1}{B} \sum_{b =1}^B \ln {P}(\mathbf{s}_b|\mathbf{x}, \theta),
\label{entropy}
\end{equation}
where $\mathbf{s}_b$ is a sampled generation to the question $\mathbf{x}$. Later \citet{kuhn2023semantic} improve the entropy by utilizing the semantics of the sampled generations. They cluster the generations with the same meaning and calculate entropy using the generation probabilities associated with each cluster: 
\begin{equation}
\mathrm{SE}(\mathbf{x},\theta)= - \frac{1}{|C|} \sum_{i=1}^{|C|} \ln P(\mathrm{c}_i|\mathbf{x}, \theta),
\label{semantic-entropy}
\end{equation}
where $\mathrm{c}_i$ refers to each semantic cluster and $C$ is the set of all clusters. Notably, \citet{aichberger2024many} enhance semantic entropy by enabling the model to generate semantically more diverse outputs.

Both \citet{malinin2021uncertainty} and \citet{kuhn2023semantic} observe that sequence probability in \eqref{normalized-prob} is biased against longer generations. To address this, they use a length-normalized scoring as follows:
\begin{equation}
     \tilde{P}(\mathbf{s}|\mathbf{x}, \theta) = \prod_{l=1}^{L}  P(s_l|s_{<l}, \mathbf{x}; \theta)^{\frac{1}{L}},
\label{length-normalized-prob}
\end{equation}
where $L$ is the sequence length.  Later \citet{bakman2024mars} and \citet{tokensar} improve this scoring function by incorporating the meaning contribution of the tokens. Their scoring functions, MARS and TokenSAR, respectively, adopt different approaches in integrating token meaning but can be generalized with the following formulation:
\begin{equation}
    \bar{P}(\mathbf{s}|\mathbf{x}, \theta) = \prod_{l=1}^{L} P(s_l|s_{<l}, \mathbf{x}; \theta)^{w(\textbf{s},\mathbf{x}, L, l)},
\label{mars}
\end{equation}
where $w(\mathbf{s}, \mathbf{x}, L, l)$ is the weight of the $l$-th token assigned by MARS or TokenSAR. These scoring functions aim to give more weight to tokens that directly answer the question and are calibrated such that if a generation is likely to be incorrect, they yield a lower score, and vice versa. Our goal in this work is to enhance this calibration by learning the scoring function directly from the data.


%

%

\section{Shortcomings of Existing Scoring Functions}\label{shortcomings}

In this section, we discuss the shortcomings of scoring functions: LNS, MARS, and TokenSAR.

\paragraph{Manually Crafted Design Choices.} Existing scoring functions are designed to address particular challenges within the UE problem domain. For instance, LNS mitigates length bias, whereas MARS and TokenSAR focus on reducing the impact of non-essential token probabilities. However, the complexities involved in designing an optimal scoring function may not be immediately apparent. Typically, scoring functions involve a dot product of log probabilities and assigned weights, but alternative formulations could provide more finely calibrated estimations. Additionally, the existing functions may not adequately capture complex dependencies between tokens, such as grammatical and semantic interactions \cite{de2019dependency}. 
While MARS attempts to address this by weighting phrases rather than individual tokens, it only partially solves the problem and might fail to capture deeper dependencies. Consider the question, "What is the tallest building in the world?" and the model's response: "The tallest building in the world might be Burj Khalifa with its lovely sight." Here, although the tokens "might" and "Burj Khalifa" may have high probabilities, "might" conveys uncertainty, suggesting that the model is uncertain despite the high probability of those tokens. An effective scoring function should recognize the interaction between "might" and "Burj Khalifa" and adjust the uncertainty accordingly. Additionally, the phrase "with its lovely sight" adds subjective opinion rather than factual reliability, yet it affects the overall meaning. Ignoring the probabilities of such tokens could improve the performance of the scoring function. Such important nuances are ignored by previous works. Lastly, both MARS and TokenSAR apply normalization on their weights $w(\mathbf{s}, \mathbf{x}, L, l)$, through methods like sum-normalization (TokenSAR) or softmax (MARS). Such design choices directly impact the UE output, potentially making the UE method converge to sub-optimal points.

\begin{figure}[]
\vskip -0.05\in
\begin{center}
\includegraphics[width=0.4\textwidth]{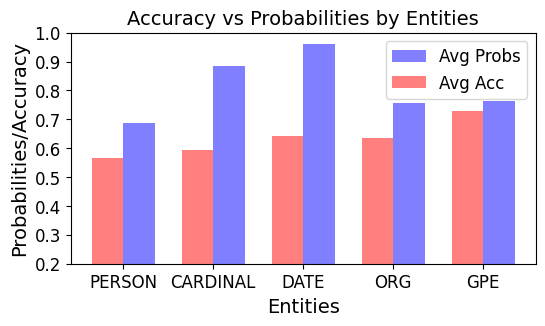}
\vskip -0.15in
\caption{Average accuracy and probability assignments of Llama2-7b-chat for specific entities in TriviaQA.}
\label{fig:entities}
\end{center}
\vskip -0.33in
\end{figure}

\paragraph{Biased Probabilities.} Existing scoring functions directly utilize token probabilities, which may be biased against certain entity types \cite{gallegos2024bias}. To explore this issue, we conducted an experiment with Llama2-7b-chat \cite{touvron2023llama2} using TriviaQA \cite{joshi2017triviaqa}. We posed TriviaQA questions to the model and analyzed the probabilities assigned to answer tokens representing different entity types like person names, organizations, and dates. Additionally, we assessed the model's accuracy on questions whose ground truth answers are in these categories. As shown in Figure \ref{fig:entities}, though the model shows comparable accuracy for date and person entities, it assigns higher probabilities to date tokens. This finding suggests a positive bias towards date entities. We observed similar trends across other entity types. These differences in probability assignments highlight the need for recalibration across entities, which current scoring functions lack.

\paragraph{Performance in Different Languages.} MARS and TokenSAR rely on existing NLP tools for implementation. Specifically, TokenSAR uses a sentence similarity model \cite{tokensar}, and MARS relies on a QA evaluator model \cite{bulian2022tomayto}. These models may not be readily available for some low-resource languages. Moreover, the design of MARS and TokenSAR is primarily oriented towards English. This orientation may be challenging applied to languages that are morphologically distinct from English. 

In the next section, we introduce a trainable scoring function, addressing these shortcomings.


\section{LARS: Learnable Response Scoring}\label{method}

\paragraph{Intuition.}We develop a new scoring function that accounts for the semantic contributions of tokens  in relation to the query, grasps biased probabilities, recognizes dependencies between tokens, and identifies other factors
that may not be immediately apparent but are crucial
for UE. Since manually designing a scoring function that has all these sophisticated properties would be extremely challenging as discussed in Section \ref{shortcomings}, we instead train  a neural network with a transformer architecture that is capable of learning these properties directly from the data. An overview of the proposed approach is visualized in Figure \ref{fig:lars}.

\paragraph{Training Strategy.} Let $f$ denote the scoring function, which accepts three arguments: the input prompt $\mathbf{x} = (x_1, x_2, \ldots, x_N)$, the generated sequence $\mathbf{s} = (s_1, s_2, \ldots, s_L)$, and the corresponding probability vector $\mathbf{p} = (p_1, p_2, \ldots, p_L)$, where $p_i$ represents the probability of token $s_i$. The function $f$ outputs a real number $o$. This mapping captures crucial information: the meaning of the generated tokens ($\mathbf{s}$), their relevance to the context provided by the input prompt ($\mathbf{x}$), and the model's confidence in each token via the probabilities ($\mathbf{p}$). In token probability-based methods, it is desirable for $o$ to be lower when the generation $\mathbf{s}$ is more likely to be incorrect, improving the model’s uncertainty estimation. To achieve this desired calibration, we make $f$ directly learnable through supervised data.

We construct a calibration set to train our scoring function, $f_w$, which is parameterized by $w$. This calibration set comprises 4-tuples: input prompt $\mathbf{x}$, generated sequence $\mathbf{s}$, probability vector $\mathbf{p}$, and binary ground truth label $g$. The label $g$ indicates whether $\mathbf{s}$ is a correct response to $\mathbf{x}$. To optimize the parameters of $f_w$, we employ the binary cross-entropy loss, denoted by $L$, applied as follows: $L(f_w(\mathbf{x}, \mathbf{s}, \mathbf{p}), g).$
To train the scoring function $f_w$, we start with the pre-trained RoBERTa-base model \cite{liu2019roberta} and augment it by adding a linear layer that outputs a single logit.

\paragraph{Input Mapping.} Inputting text sequences $x$ and $s$ into a transformer model is straightforward, as we can leverage the standard text encoding strategy \cite{vaswani2017attention}. However, encoding the probability information, which is a single real number for each token, poses a challenge due to its low dimensionality compared to the high-dimensional space of the model. To address this, we propose a novel input encoding strategy inspired by the class conditioning approach in conditional image generation \cite{NIPS2016_b1301141}. We encode probability information to high-dimensional vectors by few-hot encoding. More specifically, we partition the probability range [0,1] to $k$ partitions. These partitions are mutually exclusive, cover the entire probability range, and are determined based on the quantiles of the probabilities in the calibration dataset. Given that the transformer model has an input dimension $d$, if $p_i$ falls in the range of $r$-th partition, we set its vector positions between \((r-1) \times \frac{d}{k}\) and \(r \times \frac{d}{k}\) to 1, while all other positions are set to 0 (Figure \ref{fig:lars} Mid Right). To ensure consistency with the model's token embedding norms, we scale probability vectors by a fixed divisor and get the probability vector $\tilde{p_i}$. With this encoding strategy, we represent distinct probability ranges orthogonal to each other in high dimension. The input format of the LARS model is structured as follows (and visualized in Figure \ref{fig:lars} Mid Left): initial prompt \(\mathbf{x}\), followed by a series of response tokens \(\mathbf{s} = (s_1, s_2, \ldots, s_L)\). Each response token \(s_i\) is immediately succeeded by its probability vector $\tilde{p_i}$.

\begin{table*}[!htbp]
\centering
\vskip -0.1in
\fontsize{7.5}{5.2}\selectfont
\begin{tabular}{l|c|l| cc | cc |cc| cc|cc }
\toprule
  &UE Method & Scoring & \multicolumn{2}{c}{\textbf{Llama2-7b}} & \multicolumn{2}{c}{\textbf{Llama3-8b}} & \multicolumn{2}{c}{\textbf{Mistral-7b}}& \multicolumn{2}{c}{\textbf{Gemma-7b}} & \multicolumn{2}{c}{\textbf{Llama2-13b}} \\
   && Function & AUROC & PRR & AUROC & PRR & AUROC & PRR & AUROC & PRR & AUROC & PRR \\
\midrule[\heavyrulewidth]
\multirow{16}{*}{\rotatebox{90}{\textbf{TriviaQA}}}
&\textbf{Lex. Sim.}          &-&0.647 &0.374	&0.683 &0.483	&0.720 &0.517	&0.597 &0.227 &0.611 &0.314  \\
&\textbf{\# Sem. Gr.}     &-&0.792 &0.571	&0.819 &0.671	&0.757 &0.521	&0.744 &0.454 &0.776 &0.557\\
&$\mathbf{p}$\textbf{(True)} &-&0.616 &0.267	&0.842 &0.733    &0.805 &0.653	&0.517 &0.023 &0.650&0.392 \\
&\textbf{SAPLMA}             &-&0.741 &0.484 &0.736 &0.541    &0.785 &0.614	&0.658 &0.373  & 0.757& 0.594\\
&\textbf{Eccentricity}           &-&0.812 &0.629 &0.853 &0.756    &0.818 &0.664	&0.764 &0.496 &0.813 &0.633\\
&\textbf{Degree Matrix}          &-&0.812 &0.620 &0.851 &0.746    &0.820 &0.658	&0.766 &0.511 &0.817 &0.646\\
  \cmidrule{2-13}
&\multirow{4}{*}{\textbf{Confidence}}
&LNS         &0.697 & 0.481	&0.748 &0.600	&0.722 &0.533	&0.628 &0.281  &0.655 &0.389\\
&&MARS       &0.751 &0.576	&0.799 &0.676	&0.745 &0.593	&0.638 &0.305  &0.641 &0.381 \\
&&TokenSAR   &0.747 &0.572	&0.792 &0.674	&0.747 &0.584	&0.688 &0.386  &0.728 &0.527 \\
&&LARS       &\textbf{0.851} &\textbf{0.760}	&\textbf{0.872} &\textbf{0.817}	&\textbf{0.844} &\textbf{0.759}	&\textbf{0.819} &\textbf{0.690}  &\textbf{0.846} &\textbf{0.766} \\
  \cmidrule{2-13}
&\multirow{4}{*}{\textbf{SE}}
&LNS         &0.795 &0.627	&0.835 &0.733	&0.810 &0.670	&0.749 &0.475  &0.800 &0.617 \\
&&MARS       &0.797 &0.645	&0.835 &0.742	&0.810 &0.681	&0.749 &0.482  &0.794 &0.615 \\
&&TokenSAR   &0.796 &0.640	&0.839 &0.747	&0.813 &0.681	&0.753 &0.493  &0.806 &0.639 \\
&&LARS       &\textbf{0.849} &\textbf{0.745}	&\textbf{0.866} &\textbf{0.811}	&\textbf{0.854} &\textbf{0.782}	&\textbf{0.821} &\textbf{0.699}  &\textbf{0.866} &\textbf{0.797} \\
\midrule[\heavyrulewidth]
\multirow{16}{*}{\rotatebox{90}{\textbf{NaturalQA}}}
& \textbf{Lex. Sim.}   &-&0.600 &0.263	&0.651 &0.373	&0.637 &0.340	&0.585 &0.163 &0.604 &0.261\\
&\textbf{\# Sem. Gr.}   &-&0.705 &0.379	&0.736 &0.448	&0.675 &0.283	&0.686 &0.276 &0.709 &0.377\\
&$\mathbf{p}$\textbf{(True)}            &-&0.561 &0.090	&0.761 &0.561  &0.727 &0.509	&0.647 &0.247  &0.562 &0.131 \\
&\textbf{SAPLMA}          &-&0.691 &0.397	&0.713 &0.443 &0.723 &0.458	&0.657 &0.289  & 0.594&0.410\\
&\textbf{Eccentricity}             &-&0.727 &0.431 &0.775 &0.567    &0.727 &0.480	&0.713 &0.368 &0.741 &0.482\\
&\textbf{Degree Matrix}             &-&0.727 &0.435 &0.771 &0.554    &0.732 &0.483	&0.715 &0.358 &0.742 &0.487\\
  \cmidrule{2-13}
&\multirow{4}{*}{\textbf{Confidence}}
&LNS         &0.677 &0.384	&0.697 &0.449	&0.666 &0.390	&0.610 &0.189  &0.648 &0.338 \\
&&MARS       &0.699 &0.411	&0.717 &0.477	&0.678 &0.407	&0.615 &0.198  &0.631 &0.311 \\
&&TokenSAR   &0.703 &0.431	&0.717 &0.476	&0.682 &0.426	&0.643 &0.249  &0.677 &0.393 \\
&&LARS       &\textbf{0.780} &\textbf{0.581}	&\textbf{0.812} &\textbf{0.654}	&\textbf{0.782} &\textbf{0.599}	&\textbf{0.794} &\textbf{0.541}  &\textbf{0.772} &\textbf{0.574} \\
\cmidrule{2-13}
&\multirow{4}{*}{\textbf{SE}}
&LNS         &0.721 &0.432	&0.759 &0.548	&0.727 &0.499	&0.700 &0.332  &0.733 &0.471 \\
&&MARS       &0.720 &0.440	&0.750 &0.546	&0.725 &0.493	&0.705 &0.336  &0.723 &0.440 \\
&&TokenSAR   &0.721 &0.443	&0.756 &0.544	&0.726 &0.498	&0.700 &0.340  &0.736 &0.485 \\
&&LARS       &\textbf{0.772} &\textbf{0.569}	&\textbf{0.794} &\textbf{0.638}	&\textbf{0.778} &\textbf{0.591}	&\textbf{0.785} &\textbf{0.548}  &\textbf{0.779} &\textbf{0.583}\\

\bottomrule
\end{tabular}
\vskip -0.1in
\caption{AUROC and PRR scores of UE methods on TriviaQA, NaturalQA.}
\vskip -0.25in
\label{tab:main_results}
\end{table*}

\section{Experiments}

\subsection{Experimental Setup}
\paragraph{Test Datasets.}

To evaluate UE methods, we use a mathematical reasoning dataset and three closed-ended QA datasets. Specifically, we utilize the complete test set of GSM8K for mathematical reasoning \cite{cobbe2021gsm8k}. Following \citep{kuhn2023semantic}, we select a subset of the validation set from TriviaQA \cite{joshi2017triviaqa}. Additionally, we evaluate using the entire validation split of NaturalQA \cite{kwiatkowski2019naturalqa}. Finally, we combine the training and validation splits of Web Questions (WebQA) \cite{WebQA21}.

\paragraph{Models.} We test UE methods on 5 popular open-weight models. Llama2-7b-chat, Llama2-13b-chat \cite{touvron2023llama2} and Llama3-8b-instruct \cite{llama3modelcard} are optimized for dialogue use cases. Mistral-7b-instruct \cite{jiang2023mistral} and Gemma-7b-it \cite{gemmateam2024gemma} are instruction tuned versions of the corresponding base models. For the sake of simplicity, we drop instruction indicator words such as "-chat" in the rest of the paper.

\paragraph{Metrics.}Following previous works, we set the model's golden (most-probable) generation's correctness\footnote{With given ground truth and model generation, we use GPT-3.5-turbo for evaluating the correctness of the generation \cite{lin2023generating, tokensar, bakman2024mars}} as labels (0 and 1) and UE scores as predictions \cite{kuhn2023semantic, bakman2024mars, tokensar}. Using this, we calculate the AUROC (Area Under the Receiver Operating Characteristic), a common metric for binary classifiers \cite{kuhn2023semantic, tokensar, lin2023generating}. Since AUROC is sensitive to data imbalance, we also include the Prediction Rejection Ratio (PRR) \cite{malinin2021uncertainty}. AUROC scores range from 0.5 (random) to 1.0 (perfect), and PRR ranges from 0.0 (random) to 1.0 (perfect).

\paragraph{Baselines.}We use 4 probability-based UE methods. \textbf{Confidence} is calculated as the negative of the most likely generation's score for a given question. The other UE methods are \textbf{Entropy} as in \eqref{entropy}, \textbf{Semantic Entropy (SE)} as in \eqref{semantic-entropy}, and \textbf{SentSAR} \cite{tokensar}. Each method employs a scoring function to assign a score to a generation. We compare LARS with 3 SOTA scoring functions for this purpose: \textbf{Length-Normalized Scoring (LNS) }\cite{malinin2021uncertainty}, \textbf{MARS} \cite{bakman2024mars} and \textbf{TokenSAR} \cite{tokensar}. LARS is evaluated against these scoring functions across all probability-based UE methods. It is worth noting that combining SentSAR and TokenSAR results in the SAR method \cite{tokensar}.

\begin{table*}[!htbp]
\centering
\vskip -0.05in
\fontsize{7.2}{6}\selectfont
\begin{tabular}{c|c|l| cc |cc|cc|cc|cc }
\toprule
  &UE Method & Scoring & \multicolumn{2}{c}{\textbf{Llama2-7b}} & \multicolumn{2}{c}{\textbf{Llama3-8b}} & \multicolumn{2}{c}{\textbf{Mistral-7b}}& \multicolumn{2}{c}{\textbf{Gemma-7b}} & \multicolumn{2}{c}{\textbf{Llama2-13b}} \\
   && Function & AUROC & PRR & AUROC & PRR & AUROC & PRR & AUROC & PRR & AUROC & PRR \\
\midrule[\heavyrulewidth]
\multirow{17}{*}{\rotatebox{90}{\textbf{WebQA}}}
&\textbf{Lex. Sim.}   & -  &0.643 &0.310	&0.640 &0.321	&0.645 &0.312	&0.608 &0.214 &0.624  &0.261	  \\
&\textbf{\# Sem. Gr.} & -  &0.612 &0.138	&0.599 &0.143	&0.601 &0.184	&0.630 &0.213 &0.587  &0.157		  \\
&$\mathbf{p}$\textbf{(True)}  &-&0.558 &0.078	&0.636 &0.290	  &0.667 &0.358	  &0.552 &0.041	&0.580 &0.171	\\
&\textbf{Eccentricity} & -      &0.680 &0.375 &0.674 &0.386	&0.662 &0.333	&0.606 &0.203	&0.686  &0.358	  \\
&\textbf{Degree Matrix} & -  &0.683 &0.380 &0.676 &0.384	&0.662 &0.326	&0.611 &0.195	&0.682  &0.364	  \\
\cmidrule{2-13}
&\multirow{4}{*}{\textbf{Confidence}}
&LNS        &0.656 &0.329	&0.645 &0.324		&0.634 &0.305		&0.625 &0.246	&0.602  &0.233	 \\
&&MARS       &0.669 &0.349	&0.649 &0.333		&0.637 &0.316		&0.627 &0.258	&0.585  &0.199	 \\
&&TokenSAR   &0.664 &0.345	&0.656 &0.347		&0.640 &0.320		&0.657 &0.287	&0.615  &0.248	  \\
&&LARS (OOD) &\textbf{0.715} &\textbf{0.430}		&\textbf{0.713} &\textbf{0.464}		&\textbf{0.686} &\textbf{0.406}		&\textbf{0.726} &\textbf{0.442}	&\textbf{0.676}  &\textbf{0.367}	  \\
\cmidrule{2-13}
&\multirow{4}{*}{\textbf{SE}}
&LNS        &0.672 &0.360		&0.664 &0.366	&0.665 &0.353		&0.675 &0.334	&0.644  &0.297	 \\
&&MARS       &0.679 &0.367		&0.667 &0.370	&0.665 &0.354		&0.679 &0.340	 &0.632  &0.267	 \\
&&TokenSAR   &0.674 &0.365		&0.667 &0.372	&0.663 &0.351		&0.680 &0.343	 &0.647  &0.298	 \\
&&LARS (OOD) &\textbf{0.711} &\textbf{0.440}		&\textbf{0.694} &\textbf{0.449}		&\textbf{0.697} &\textbf{0.430}		&\textbf{0.719} &\textbf{0.440}	&\textbf{0.678}  &\textbf{0.382}	 \\

\midrule[\heavyrulewidth]
\multirow{19}{*}{\rotatebox{90}{\textbf{GSM8K}}}
&\textbf{Lex. Sim.}	&-	&0.444	&0.000	&0.632	&0.272	&0.537	&0.019	&0.544	&0.080	&0.551	&0.110	\\
&\textbf{\# Sem. Gr.}	&-	&0.513	&0.000	&0.584	&0.138	&0.532	&0.037	&0.566	&0.114	&0.561	&0.065	\\
&$\mathbf{p}$\textbf{(True)}	&-	&0.540	&0.099	&0.797	&0.623	&0.665	&0.238	&0.486	&0.000	&0.501	&0.000	\\
&\textbf{Eccentricity}	&-	&0.547	&0.049	&0.664	&0.384	&0.584	&0.109	&0.595	&0.146	&0.600	&0.163	\\
&\textbf{Degree Matrix}	&-	&0.535	&0.056	&0.667	&0.667	&0.604	&0.165	&0.584	&0.117	&0.605	&0.179	\\

\cmidrule{2-13}
&\multirow{5}{*}{\textbf{Confidence}}
&LNS	&0.570	&0.031	&0.686	&0.390	&0.567	&0.072	&0.556	&0.370	&0.615	&0.196	\\
	&&MARS	&0.567	&0.010	&0.713	&0.438	&0.568	&0.076	&0.541	&0.099	&0.562	&0.114	\\
	&&TokenSAR	&0.579	&0.045	&0.719	&0.460	&0.619	&0.156	&0.579	&0.161	&0.636	&0.233	\\
	&&LARS	&\textbf{0.720}	&\textbf{0.319}	&\textbf{0.836}	&\textbf{0.711}	&\textbf{0.708}	&\textbf{0.350}	&\textbf{0.706}	&\textbf{0.370}	&\textbf{0.738}	&\textbf{0.497}	\\
	&&LARS (OOD)	&0.603	&0.097	&0.684	&0.348	&0.630	&0.188	&0.576	&0.114	&0.635	&0.218	\\

\cmidrule{2-13}
&\multirow{5}{*}{\textbf{SE}}
&LNS	&0.516	&0.000	&0.633	&0.321	&0.560	&0.076	&0.588	&0.141	&0.587	&0.153	\\
&&MARS	&0.513	&0.000	&0.640	&0.344	&0.563	&0.080	&0.586	&0.134	&0.583	&0.122	\\
&&TokenSAR	&0.526	&0.005	&0.638	&0.344	&0.578	&0.102	&0.588	&0.148	&0.592	&0.171	\\
&&LARS	&\textbf{0.675}	&\textbf{0.267}	&\textbf{0.715}	&\textbf{0.528}	&\textbf{0.663}	&\textbf{0.310}	&\textbf{0.679}	&\textbf{0.345}	&\textbf{0.697}	&\textbf{0.383}	\\
&&LARS (OOD)	&0.572	&0.072	&0.633	&0.298	&0.605	&0.170	&0.579	&0.112	&0.608	&0.209	\\
\bottomrule

\end{tabular}
\vskip -0.1in
\caption{AUROC and PRR scores of UE methods on WebQA and GSM8K. LARS (OOD) denotes that the LARS model is trained with TriviaQA and NaturalQA.}
\vskip -0.25in
\label{tab:webqa}
\end{table*}

Further, we add 6 non-probability-based UE approaches to our baseline set. \textbf{Lexical Similarity} \cite{lexical_sim}, is the average of the Rouge-L scores between unique sampled generation pairs to a given question. $\mathbf{p}$\textbf{(True)} \cite{kadavath2022language}, a self-check method, asks the model itself if the most likely answer is correct by providing the question, sampled generations, and the answer. \textbf{SAPLMA} \cite{azaria-mitchell-2023-internal} is a probing-based method that trains the model's internal representations to predict the correctness of its generation. \textbf{Eccentricity and Degree Matrix} \cite{lin2023generating} assesses output consistency using different linear algebraic techniques. Lastly, \# \textbf{Semantic Groups} \cite{kuhn2023semantic} is the number of semantic clusters, as in SE. In all of our experiments, number of sampled generations is 5. 

\paragraph{LARS Calibration Datasets.} To train the model of the proposed method LARS, we employ train splits of TriviaQA, NaturalQA, and GSM8K. We sample six generations per question, ensuring the most likely generation is included, for each aforementioned model. From these generations, we curate unique QA pairs for calibration data and use GPT-3.5-turbo to evaluate their correctness. Typically, we train distinct LARS models for each model-dataset combination.   In some experiments, we merge TriviaQA and NaturalQA per model and train accordingly, which we specify when used. Further details are presented in Appendix \ref{exp_details}.

\subsection{Main Results}

We present the results of our method alongside other baselines in Table \ref{tab:main_results} and extended results in Appendix \ref{additional_experiments}. Notably, LARS significantly enhances the performance of all existing scoring functions across each probability-based UE method, with improvements reaching up to 0.231 AUROC and 0.46 PRR points over LNS. Additionally, LARS boosts the confidence metric to levels comparable with SE. This is particularly important considering the inference costs. Entropy-based methods require multiple output samples (5 in our experiments), which can be computationally expensive in the context of LLMs. Further, SE requires $O(N^2)$ model passes for semantic clustering, where $N$ is the number of sampled outputs. In contrast, LARS operates with a single pass using a RoBERTa-based model with 125M parameters—a computation level that is negligible compared to models with capacities of 7B parameters or more. Notably, LARS outperforms SAPLMA, which also uses the same amount of supervised data. Additionally, LARS consistently surpasses response clustering methods that require multiple output samples, such as Lexical Similarity, the Number of Semantic Groups, Eccentricity, Degree Matrix, and p(True) method.

\subsection{Out-of-Distribution (OOD)
Experiments}

We train LARS using a calibration dataset, which is curated from a set of questions and the corresponding responses of a chat model. It is crucial to assess the out-of-distribution capabilities of LARS, which we analyze from two perspectives in this section.

\paragraph{OOD Data Generalization.} First, we investigate how the performance of LARS is affected when the model encounters questions which have a distribution deviating from that of the calibration set. To this end, we conduct tests using WebQA and GSM8K, with LARS models trained on combined TriviaQA and NaturalQA for each distinct chat model. The results are presented in Table \ref{tab:webqa}, and additional results on out-of-distribution (OOD) data generalization are available in Appendix \ref{app:ood_data}. Impressively, LARS, despite being trained on different datasets, outperforms all other scoring functions across all probability-based UE methods in WebQA, achieving an average improvement of approximately $0.04$ AUROC points. However, in the GSM8K dataset, where the model was trained on a different task, performance degradation becomes significant, highlighting the importance of training LARS on task-specific data for optimal results. This performance gap may be attributed to differences in the nature of the datasets: while TriviaQA answers are primarily composed of entities such as person and organization names, GSM8K primarily involves numerical answers. As a result, calibrating LARS for entity-based answers in TriviaQA makes it less effective for GSM8K, compared to direct calibration on GSM8K itself. Nevertheless, LARS still outperforms other scoring functions in all models except Llama-3-8b, even when not specifically calibrated for the correct dataset.

\paragraph{OOD Model Generalization.} Next, we analyze how LARS performs when the responses in the calibration set are derived from a different chat model than the one used at test time. Due to space limitations, we provide a subset of the results in Table~\ref{tab:ood_model_small}; however, comprehensive results are presented in Appendix~\ref{app:ood_model}. Optimal LARS performance is achieved when the same chat model is used for both training and testing. Nevertheless, OOD model scores still surpass those of baseline scoring functions (see Tables \ref{tab:main_results} and \ref{tab:main_results_big} for baselines), confirming the effectiveness of LARS.

\begin{table}[!htbp]
\centering
\vskip -0.1in
\fontsize{8.2}{8}\selectfont
\begin{tabular}{c|l| ccc }
\toprule
\thead{UE\\Method}& {\thead{Calib\\Model}} & \textbf{\thead{Llama2\\7b}} & \textbf{\thead{Llama3\\8b}} & \textbf{\thead{Mistral\\7b}}\\
\midrule[\heavyrulewidth]
\multirow{3}{*}{\textbf{Confidence}}
&Llama2-7b       &\textbf{0.858}	&0.836	&0.831	  \\
&Llama3-8b       &0.852	&\textbf{0.874}	&0.850	  \\
&Mistral-7b      &0.835	&0.833	&\textbf{0.852}	 \\
\midrule[\heavyrulewidth]
\multirow{3}{*}{\textbf{Entropy}}
&Llama2-7b       &{0.847}	&0.830	&0.827	 \\
&Llama3-8b       &\textbf{0.852} &\textbf{0.873}	&0.850	\\
&Mistral-7b      &0.841	&0.841	&\textbf{0.854} \\
\midrule[\heavyrulewidth]
\multirow{3}{*}{\textbf{SE}}
&Llama2-7b       &{0.850}	&0.836	&0.840	  \\
&Llama3-8b       &\textbf{0.863}	&\textbf{0.872}	&\textbf{0.862}	 \\
&Mistral-7b      &0.850	&0.849	&{0.859}  \\
\midrule[\heavyrulewidth]
\multirow{3}{*}{\textbf{SentSAR}}
&Llama2-7b       &0.857	&0.841	&0.841	  \\
&Llama3-8b       &\textbf{0.866}	&\textbf{0.884}	&\textbf{0.863}	  \\
&Mistral-7b      &0.851	&0.847	&{0.860}	  \\
\bottomrule
\end{tabular}
\vskip -0.1in
\caption{AUROC scores of UE methods with LARS models trained with answers from various chat models.}
\vskip -0.2in
\label{tab:ood_model_small}
\end{table}

\subsection{LARS on Different Languages}\label{turkish}
To evaluate the performance of LARS and other scoring functions across different languages, we translated the TriviaQA test and calibration datasets into Turkish, German, and Spanish. As shown in Table \ref{tab:main_results_lang}, LARS demonstrates adaptability across languages and outperforms existing scoring functions, showing the importance of calibrating scoring functions for multilingual applications.


\begin{table}[!htbp]
\vskip -0.1in
\centering
\fontsize{8.5}{8}\selectfont
\begin{tabular}{l| cccc }
\toprule
Scoring Func. & \textbf{English} & \textbf{Turkish}  & \textbf{German} & \textbf{Spanish}\\
\midrule[\heavyrulewidth]
LNS        &0.747  &0.692 &0.710 &0.701  \\
MARS       &0.801  &0.695 &0.728 &0.723 \\
TokenSAR   &0.793  &0.720 &0.758 &0.750 \\
LARS       &\textbf{0.864} &\textbf{0.814}& \textbf{0.827} & \textbf{0.835}\\
\bottomrule
\end{tabular}
\vskip -0.1in
\caption{AUROC performance of Entropy with different scoring functions on Llama3-8B for the TriviaQA dataset in different languages.}
\vskip -0.25in
\label{tab:main_results_lang}
\end{table}

\section{Ablation Studies}

\subsection{Probability Association Strategies}\label{sec:v1-v2}
\begin{figure}[!htbp]
\vskip -0.1in
\begin{center}
\includegraphics[width=0.49\textwidth]{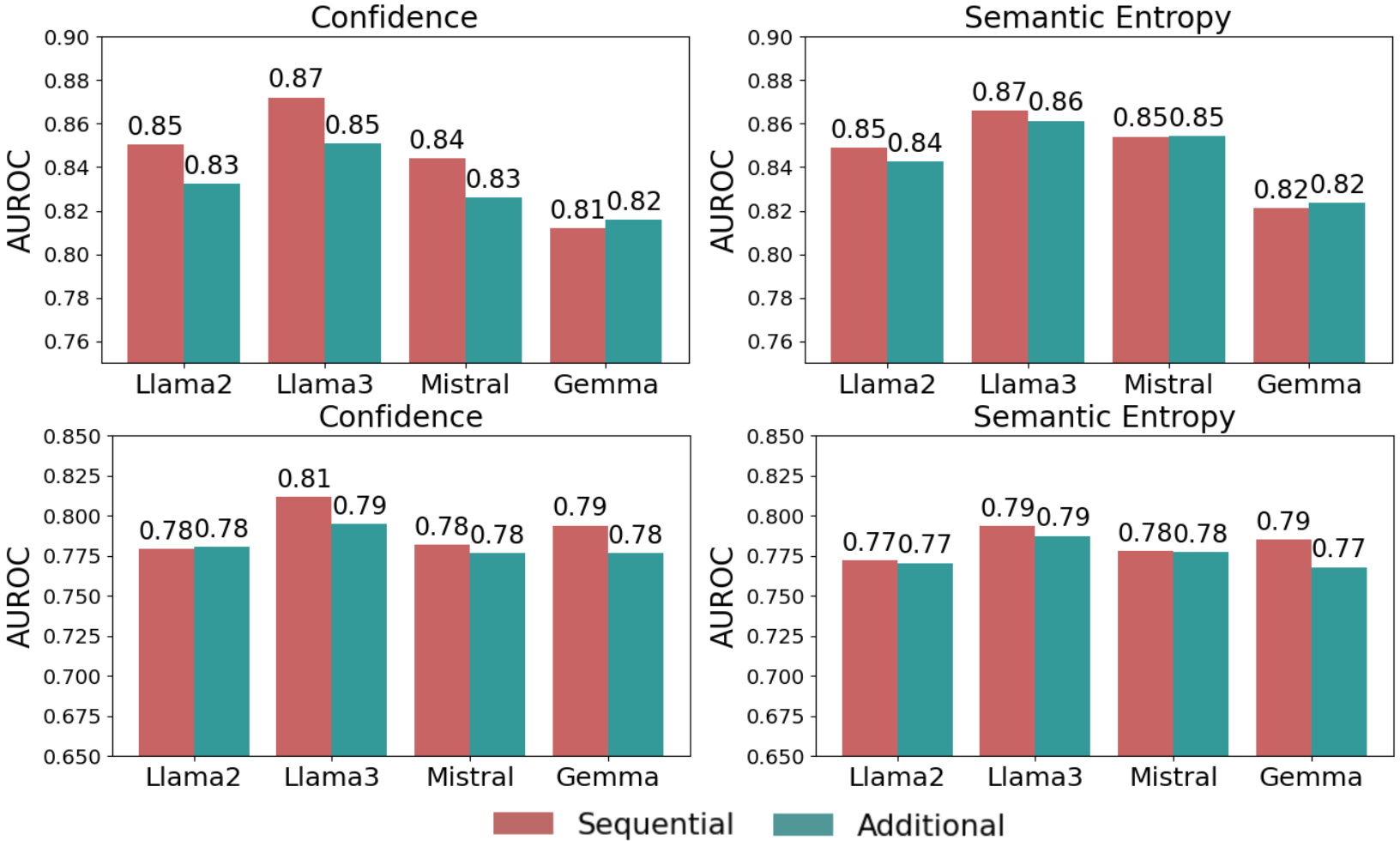}
\vskip -0.12in
\caption{Comparison of different probability association methods for LARS on TriviaQA (top) and NaturalQA (bottom).}
\label{fig:v1_v2_small}
\end{center}
\vskip -0.3in
\end{figure}

In Section \ref{method}, we explain a sequential approach to associate tokens of the response with their probabilities, where probability vectors are placed after each response token in the input to LARS. As an alternative, we explore an additive approach. In this method, the embedding vectors of the probabilities are added to the embedding vectors of their corresponding response tokens. This strategy effectively reduces the input sequence length for the LARS model. 
Results in Figure \ref{fig:v1_v2_small} demonstrate that the sequential approach is, on average, 0.15 points better when used with Confidence, although the gap narrows for SE. Comparing the additive approach with other baselines from Table \ref{tab:main_results}, we observe that it still significantly outperforms the baselines. Overall, these two probability association approaches highlight a possible trade-off between shortened input length (to the LARS model) and improved UE performance. Extended results for this experiment are presented in Appendix \ref{app:prob_asc}.

\subsection{Size of the Calibration Dataset} 
\begin{figure}[!htbp]
\vskip -0.1in
\begin{center}
\includegraphics[width=0.47\textwidth]{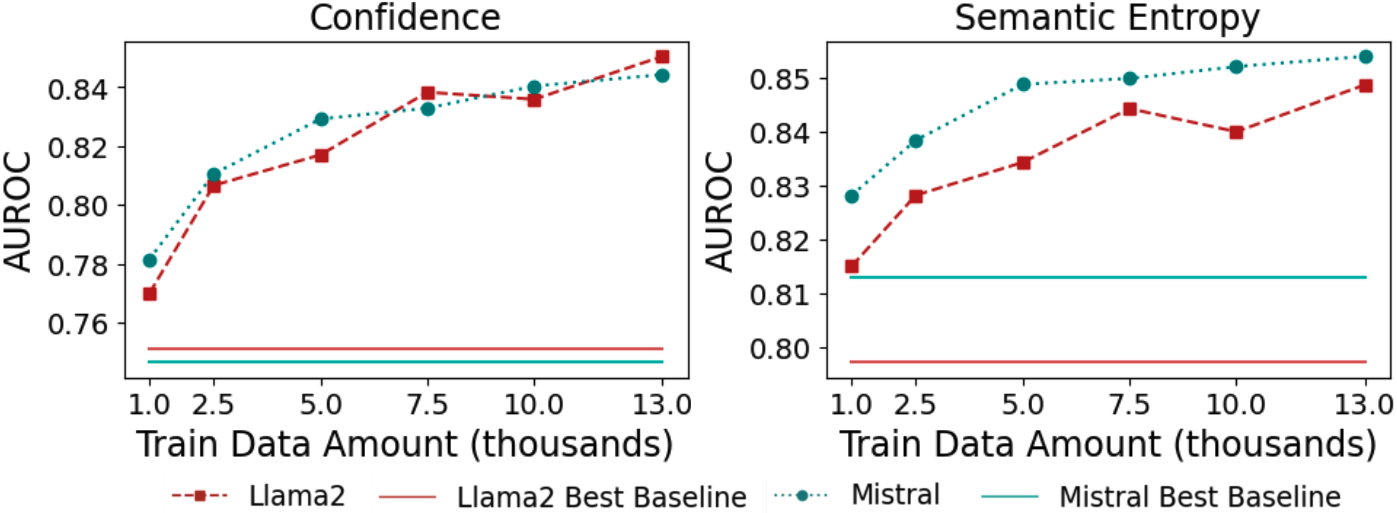}
\vskip -0.1in
\caption{AUROC scores of LARS for different amount of questions in calibration data on TriviaQA. For each UE method and model, the best score across baseline scoring functions is provided as a reference. }
\label{fig:data_amount_small}
\end{center}
\vskip -0.3in
\end{figure}

To assess the scalability of LARS, we calibrate it using varying amounts of labeled data. Results in Figure \ref{fig:data_amount_small} show that even with as few as 1,000 labeled question-ground truth pairs, LARS outperforms the best-performing baseline. Impressively, LARS demonstrates good scalability with calibration data size. Exploring the scaling of LARS with even more data remains as a future direction.

\subsection{Importance of LARS Input Components}\label{lars_components}

\paragraph{Effect of Probability Information.}
To assess the importance of probability information for LARS, we train a version of the model using only textual inputs as: the question and the generated answer \citet{feng-etal-2024-dont} did. The results (Table \ref{tab:no_prob}) indicate that excluding probability information leads to a decrease in the performance of LARS by up to 0.101 AUROC score. This shows the critical role of probability information in LARS.

\paragraph{Effect of Textual Information.}

To assess the impact of textual and semantic information in the input, we conduct an experiment using only the probability information. Specifically, we train a Multilayer Perceptron (MLP) with two hidden layers, which accepts only the probability vector as input. As presented in Table~\ref{tab:no_prob}, the probability-only model achieves an AUROC of \textbf{0.721} with the Confidence metric, significantly underperforming compared to MARS \textbf{(0.751)}, TokenSAR \textbf{(0.747)}, and LARS \textbf{(0.851)}. These results highlight the crucial role of integrating textual and probability information in enhancing the performance of LARS.

\begin{table}[!htbp]
\vskip -0.1in
\centering
\fontsize{8.5}{7.5}\selectfont
\begin{tabular}{c|l| ccc }
\toprule
{UE Method}& {Scoring Function} & {AUROC} & PRR\\
\midrule[\heavyrulewidth]
\multirow{4}{*}{\textbf{Confidence}}
&Only text       &0.750	&0.581  \\
&Only probs      &0.721 & 0.372\\
&LARS            &\textbf{0.851} & \textbf{0.760}\\
\midrule[\heavyrulewidth]
\multirow{4}{*}{\textbf{Entropy}}
&Only text       &0.754 &0.592 	  \\
&Only probs      &0.733 & 0.507 \\
&LARS            &\textbf{0.842} &\textbf{0.748} \\
\midrule[\heavyrulewidth]
\multirow{4}{*}{\textbf{SE}}
&Only text       &0.817 &0.711   \\
&Only probs      &0.799 &   0.623\\
&LARS            &\textbf{0.849} &\textbf{0.745} 	\\
\midrule[\heavyrulewidth]
\multirow{4}{*}{\textbf{SentSAR}}
&Only text       &0.783   &0.664 \\
&Only probs      & 0.771 & 0.589 \\
&LARS            &\textbf{0.850} &\textbf{0.763} 	\\
\bottomrule
\end{tabular}
\vskip -0.1in
\caption{ Comparison of different input modalities (text-only, probabilities-only, and combined text and probabilities) with Llama2-7b model on the TriviaQA.}

\vskip -0.2in
\label{tab:no_prob}
\end{table}

\section{Related Works}\label{related-work}

 UE has recently become a topic of significant interest, leading to the proposal of various methods. These methods can be broadly categorized into four types: 1. Self-checking methods: The model evaluates its own generation correctness using different strategies \cite{kadavath2022language, manakul2023selfcheckgpt,li2024think,luo2023zeroresource,zhao2023knowing}. 2. Output consistency methods: Uncertainty is predicted by examining the consistency of various outputs for a given question \cite{lyu2024calibrating, lin2023generating,zhang2024sac3,ulmer2024calibrating,elaraby2023halo}. 3. Internal state examination methods: The activations of the model are analyzed to predict the model errors \cite{chen2024inside}. 4. Token probability-based methods: Token probabilities are utilized to estimate uncertainty \cite{malinin2021uncertainty, kuhn2023semantic, bakman2024mars, tokensar}. 
 

Several approaches \cite{lu-etal-2022-learning, ravi2024lynxopensourcehallucination, azaria-mitchell-2023-internal, feng-etal-2024-dont} have utilized supervised training to predict model generation reliability in various contexts, such as hallucination detection and machine translation. \citet{lu-etal-2022-learning, azaria-mitchell-2023-internal} trained simple neural networks that take an internal state as input and output generation correctness. From a practical perspective, this approach has limitations compared to LARS, as accessing model activations is not feasible for closed-weight models. Additionally, using internal states might not be ideal for predicting correctness, since these states contain diverse information which may be irrelevant for assessing reliability \cite{huben2024sparse}. In Table \ref{tab:main_results}, we demonstrate that LARS significantly outperforms the approach of \citet{azaria-mitchell-2023-internal}. Moreover, selecting which internal state to use remains an open question, as the optimal state can vary from model to model. Transferability across models is also constrained, particularly when dealing with differing internal dimensions, whereas LARS exhibits strong model-transferability performance. 
Another line of work by \citet{ravi2024lynxopensourcehallucination} trains a separate generative LLM (observer LLM) using input and corrected output pairs along with the reasoning for corrections to detect errors in the generation. Observer LLM relies on its own reasoning and general knowledge capabilities to detect hallucinations. Overall, this method requires fine-tuning of a generative pretrained LLM with big sizes such as 70B parameters and high-quality data curated by human experts. Conversely, LARS uses the model's probabilities and the generation to calibrate the UE computation. Therefore, our approach does not require training a very large generative model unlike \citet{ravi2024lynxopensourcehallucination} because LARS does not rely on model's own factual knowledge and reasoning capabilities. Their approach can be adapted to our setting by training only question-text pairs with RoBERTa model which performs poorly compared to LARS as shown in Section \ref{lars_components}.

\section{Conclusion}
In this study, we first demonstrated the shortcomings of existing scoring functions for uncertainty estimation in LLMs. Then, we introduced LARS, an off-the-shelf scoring function directly learned from data. We demonstrated that LARS significantly outperforms existing baselines across three different QA datasets, a mathematical reasoning task, and four different languages. Further, our results indicate that LARS' performance scales well with increased data.

\section{Limitations}

One limitation of LARS is its reliance on labeled data, which is not a requirement for other scoring functions. Further, LARS depends on a pretrained RoBERTa model, which has a limited sequence length capability. This may necessitate the pre-training of BERT-like models that can handle longer sequences. Lastly, training LARS with a transformer model reduces the interpretability of the features. Traditional scoring functions modify the weighting of probabilities and compute a dot product between log probabilities and weights, offering a level of interpretability. LARS, however, lacks it due to being a more complex function (despite its superior performance).

\section{Ethics Statement}
Although LARS demonstrates superior performance compared to existing scoring functions, it is important to remember that these methods still fall short of perfection. Consequently, the results from UE methods should still be taken with a grain of salt, especially in critical domains such as law and medicine. Additionally, LARS may propagate any biases that may be present in its training data into the scoring function, potentially introducing biases in UE related to gender, ethnicity, age, and so on. Such risks must be carefully managed in real-world applications.

\bibliography{custom}

\begin{thebibliography}{46}
\expandafter\ifx\csname natexlab\endcsname\relax\def\natexlab#1{#1}\fi

\bibitem[{Aichberger et~al.(2024)Aichberger, Schweighofer, Ielanskyi, and Hochreiter}]{aichberger2024many}
Lukas Aichberger, Kajetan Schweighofer, Mykyta Ielanskyi, and Sepp Hochreiter. 2024.
\newblock How many opinions does your llm have? improving uncertainty estimation in nlg.
\newblock In \emph{ICLR 2024 Workshop on Secure and Trustworthy Large Language Models}.

\bibitem[{AI@Meta(2024)}]{llama3modelcard}
AI@Meta. 2024.
\newblock \href {https://github.com/meta-llama/llama3/blob/main/MODEL_CARD.md} {Llama 3 model card}.

\bibitem[{Azaria and Mitchell(2023)}]{azaria-mitchell-2023-internal}
Amos Azaria and Tom Mitchell. 2023.
\newblock \href {https://doi.org/10.18653/v1/2023.findings-emnlp.68} {The internal state of an {LLM} knows when it{'}s lying}.
\newblock In \emph{Findings of the Association for Computational Linguistics: EMNLP 2023}, pages 967--976, Singapore. Association for Computational Linguistics.

\bibitem[{Bakman et~al.(2024)Bakman, Yaldiz, Buyukates, Tao, Dimitriadis, and Avestimehr}]{bakman2024mars}
Yavuz~Faruk Bakman, Duygu~Nur Yaldiz, Baturalp Buyukates, Chenyang Tao, Dimitrios Dimitriadis, and Salman Avestimehr. 2024.
\newblock \href {https://doi.org/10.18653/v1/2024.acl-long.419} {{MARS}: Meaning-aware response scoring for uncertainty estimation in generative {LLM}s}.
\newblock In \emph{Proceedings of the 62nd Annual Meeting of the Association for Computational Linguistics (Volume 1: Long Papers)}, pages 7752--7767, Bangkok, Thailand. Association for Computational Linguistics.

\bibitem[{Berant et~al.(2013)Berant, Chou, Frostig, and Liang}]{WebQA21}
Jonathan Berant, Andrew Chou, Roy Frostig, and Percy Liang. 2013.
\newblock \href {https://www.aclweb.org/anthology/D13-1160} {Semantic parsing on {F}reebase from question-answer pairs}.
\newblock In \emph{Proceedings of the 2013 Conference on Empirical Methods in Natural Language Processing}, pages 1533--1544, Seattle, Washington, USA. Association for Computational Linguistics.

\bibitem[{Bulian et~al.(2022)Bulian, Buck, Gajewski, B{\"o}rschinger, and Schuster}]{bulian2022tomayto}
Jannis Bulian, Christian Buck, Wojciech Gajewski, Benjamin B{\"o}rschinger, and Tal Schuster. 2022.
\newblock \href {https://doi.org/10.18653/v1/2022.emnlp-main.20} {Tomayto, tomahto. beyond token-level answer equivalence for question answering evaluation}.
\newblock In \emph{Proceedings of the 2022 Conference on Empirical Methods in Natural Language Processing}, pages 291--305.

\bibitem[{Chen et~al.(2024)Chen, Liu, Chen, Gu, Wu, Tao, Fu, and Ye}]{chen2024inside}
Chao Chen, Kai Liu, Ze~Chen, Yi~Gu, Yue Wu, Mingyuan Tao, Zhihang Fu, and Jieping Ye. 2024.
\newblock \href {https://openreview.net/forum?id=Zj12nzlQbz} {{INSIDE}: {LLM}s' internal states retain the power of hallucination detection}.
\newblock In \emph{The Twelfth International Conference on Learning Representations}.

\bibitem[{Chern et~al.(2023)Chern, Chern, Chen, Yuan, Feng, Zhou, He, Neubig, Liu et~al.}]{chern2023factool}
I~Chern, Steffi Chern, Shiqi Chen, Weizhe Yuan, Kehua Feng, Chunting Zhou, Junxian He, Graham Neubig, Pengfei Liu, et~al. 2023.
\newblock Factool: Factuality detection in generative ai--a tool augmented framework for multi-task and multi-domain scenarios.
\newblock \emph{arXiv preprint arXiv:2307.13528}.

\bibitem[{Cobbe et~al.(2021)Cobbe, Kosaraju, Bavarian, Chen, Jun, Kaiser, Plappert, Tworek, Hilton, Nakano, Hesse, and Schulman}]{cobbe2021gsm8k}
Karl Cobbe, Vineet Kosaraju, Mohammad Bavarian, Mark Chen, Heewoo Jun, Lukasz Kaiser, Matthias Plappert, Jerry Tworek, Jacob Hilton, Reiichiro Nakano, Christopher Hesse, and John Schulman. 2021.
\newblock Training verifiers to solve math word problems.
\newblock \emph{arXiv preprint arXiv:2110.14168}.

\bibitem[{Cohen et~al.(2023)Cohen, Hamri, Geva, and Globerson}]{cohen-etal-2023-lm}
Roi Cohen, May Hamri, Mor Geva, and Amir Globerson. 2023.
\newblock \href {https://doi.org/10.18653/v1/2023.emnlp-main.778} {{LM} vs {LM}: Detecting factual errors via cross examination}.
\newblock In \emph{Proceedings of the 2023 Conference on Empirical Methods in Natural Language Processing}, pages 12621--12640, Singapore. Association for Computational Linguistics.

\bibitem[{De~Marneffe and Nivre(2019)}]{de2019dependency}
Marie-Catherine De~Marneffe and Joakim Nivre. 2019.
\newblock Dependency grammar.
\newblock \emph{Annual Review of Linguistics}, 5:197--218.

\bibitem[{Devlin et~al.(2019)Devlin, Chang, Lee, and Toutanova}]{devlin2019bert}
Jacob Devlin, Ming-Wei Chang, Kenton Lee, and Kristina Toutanova. 2019.
\newblock \href {https://doi.org/10.18653/v1/N19-1423} {{BERT}: Pre-training of deep bidirectional transformers for language understanding}.
\newblock In \emph{Proceedings of the 2019 Conference of the North {A}merican Chapter of the Association for Computational Linguistics: Human Language Technologies, Volume 1 (Long and Short Papers)}, pages 4171--4186.

\bibitem[{Duan et~al.(2024)Duan, Cheng, Wang, Zavalny, Wang, Xu, Kailkhura, and Xu}]{tokensar}
Jinhao Duan, Hao Cheng, Shiqi Wang, Alex Zavalny, Chenan Wang, Renjing Xu, Bhavya Kailkhura, and Kaidi Xu. 2024.
\newblock \href {https://doi.org/10.18653/v1/2024.acl-long.276} {Shifting attention to relevance: Towards the predictive uncertainty quantification of free-form large language models}.
\newblock In \emph{Proceedings of the 62nd Annual Meeting of the Association for Computational Linguistics (Volume 1: Long Papers)}, pages 5050--5063, Bangkok, Thailand. Association for Computational Linguistics.

\bibitem[{Elaraby et~al.(2023)Elaraby, Lu, Dunn, Zhang, Wang, and Liu}]{elaraby2023halo}
Mohamed Elaraby, Mengyin Lu, Jacob Dunn, Xueying Zhang, Yu~Wang, and Shizhu Liu. 2023.
\newblock Halo: Estimation and reduction of hallucinations in open-source weak large language models.
\newblock \emph{arXiv preprint arXiv:2308.11764}.

\bibitem[{Feng et~al.(2024)Feng, Shi, Wang, Ding, Balachandran, and Tsvetkov}]{feng-etal-2024-dont}
Shangbin Feng, Weijia Shi, Yike Wang, Wenxuan Ding, Vidhisha Balachandran, and Yulia Tsvetkov. 2024.
\newblock \href {https://doi.org/10.18653/v1/2024.acl-long.786} {Don{'}t hallucinate, abstain: Identifying {LLM} knowledge gaps via multi-{LLM} collaboration}.
\newblock In \emph{Proceedings of the 62nd Annual Meeting of the Association for Computational Linguistics (Volume 1: Long Papers)}, pages 14664--14690, Bangkok, Thailand. Association for Computational Linguistics.

\bibitem[{Fomicheva et~al.(2020)Fomicheva, Sun, Yankovskaya, Blain, Guzm{\'a}n, Fishel, Aletras, Chaudhary, and Specia}]{lexical_sim}
Marina Fomicheva, Shuo Sun, Lisa Yankovskaya, Fr{\'e}d{\'e}ric Blain, Francisco Guzm{\'a}n, Mark Fishel, Nikolaos Aletras, Vishrav Chaudhary, and Lucia Specia. 2020.
\newblock \href {https://doi.org/10.1162/tacl_a_00330} {Unsupervised quality estimation for neural machine translation}.
\newblock \emph{Transactions of the Association for Computational Linguistics}, 8:539--555.

\bibitem[{Gallegos et~al.(2024)Gallegos, Rossi, Barrow, Tanjim, Kim, Dernoncourt, Yu, Zhang, and Ahmed}]{gallegos2024bias}
Isabel~O Gallegos, Ryan~A Rossi, Joe Barrow, Md~Mehrab Tanjim, Sungchul Kim, Franck Dernoncourt, Tong Yu, Ruiyi Zhang, and Nesreen~K Ahmed. 2024.
\newblock Bias and fairness in large language models: A survey.
\newblock \emph{Computational Linguistics}, pages 1--79.

\bibitem[{Huang et~al.(2023)Huang, Vora, Liang, and Leskovec}]{huang2023benchmarking}
Qian Huang, Jian Vora, Percy Liang, and Jure Leskovec. 2023.
\newblock \href {https://openreview.net/forum?id=kXlTY0BmK3} {Benchmarking large language models as {AI} research agents}.
\newblock In \emph{NeurIPS 2023 Foundation Models for Decision Making Workshop}.

\bibitem[{Huben et~al.(2024)Huben, Cunningham, Smith, Ewart, and Sharkey}]{huben2024sparse}
Robert Huben, Hoagy Cunningham, Logan~Riggs Smith, Aidan Ewart, and Lee Sharkey. 2024.
\newblock \href {https://openreview.net/forum?id=F76bwRSLeK} {Sparse autoencoders find highly interpretable features in language models}.
\newblock In \emph{The Twelfth International Conference on Learning Representations}.

\bibitem[{Jiang et~al.(2023)Jiang, Sablayrolles, Mensch, Bamford, Chaplot, de~las Casas, Bressand, Lengyel, Lample, Saulnier, Lavaud, Lachaux, Stock, Scao, Lavril, Wang, Lacroix, and Sayed}]{jiang2023mistral}
Albert~Q. Jiang, Alexandre Sablayrolles, Arthur Mensch, Chris Bamford, Devendra~Singh Chaplot, Diego de~las Casas, Florian Bressand, Gianna Lengyel, Guillaume Lample, Lucile Saulnier, Lélio~Renard Lavaud, Marie-Anne Lachaux, Pierre Stock, Teven~Le Scao, Thibaut Lavril, Thomas Wang, Timothée Lacroix, and William~El Sayed. 2023.
\newblock \href {http://arxiv.org/abs/2310.06825} {Mistral 7b}.

\bibitem[{Joshi et~al.(2017)Joshi, Choi, Weld, and Zettlemoyer}]{joshi2017triviaqa}
Mandar Joshi, Eunsol Choi, Daniel Weld, and Luke Zettlemoyer. 2017.
\newblock \href {https://doi.org/10.18653/v1/P17-1147} {{T}rivia{QA}: A large scale distantly supervised challenge dataset for reading comprehension}.
\newblock In \emph{Proceedings of the 55th Annual Meeting of the Association for Computational Linguistics (Volume 1: Long Papers)}, pages 1601--1611, Vancouver, Canada. Association for Computational Linguistics.

\bibitem[{Kadavath et~al.(2022)Kadavath, Conerly, Askell, Henighan, Drain, Perez, Schiefer, Hatfield-Dodds, DasSarma, Tran-Johnson, Johnston, El-Showk, Jones, Elhage, Hume, Chen, Bai, Bowman, Fort, Ganguli, Hernandez, Jacobson, Kernion, Kravec, Lovitt, Ndousse, Olsson, Ringer, Amodei, Brown, Clark, Joseph, Mann, McCandlish, Olah, and Kaplan}]{kadavath2022language}
Saurav Kadavath, Tom Conerly, Amanda Askell, Tom Henighan, Dawn Drain, Ethan Perez, Nicholas Schiefer, Zac Hatfield-Dodds, Nova DasSarma, Eli Tran-Johnson, Scott Johnston, Sheer El-Showk, Andy Jones, Nelson Elhage, Tristan Hume, Anna Chen, Yuntao Bai, Sam Bowman, Stanislav Fort, Deep Ganguli, Danny Hernandez, Josh Jacobson, Jackson Kernion, Shauna Kravec, Liane Lovitt, Kamal Ndousse, Catherine Olsson, Sam Ringer, Dario Amodei, Tom Brown, Jack Clark, Nicholas Joseph, Ben Mann, Sam McCandlish, Chris Olah, and Jared Kaplan. 2022.
\newblock \href {http://arxiv.org/abs/2207.05221} {Language models (mostly) know what they know}.

\bibitem[{Kim et~al.(2021)Kim, Koo, and Kim}]{kim2021envbert}
Dohyung Kim, Jahwan Koo, and Ung-Mo Kim. 2021.
\newblock Envbert: multi-label text classification for imbalanced, noisy environmental news data.
\newblock In \emph{2021 15th International Conference on Ubiquitous Information Management and Communication (IMCOM)}, pages 1--8. IEEE.

\bibitem[{Kuhn et~al.(2023)Kuhn, Gal, and Farquhar}]{kuhn2023semantic}
Lorenz Kuhn, Yarin Gal, and Sebastian Farquhar. 2023.
\newblock \href {https://openreview.net/forum?id=VD-AYtP0dve} {Semantic uncertainty: Linguistic invariances for uncertainty estimation in natural language generation}.
\newblock In \emph{The Eleventh International Conference on Learning Representations}.

\bibitem[{Kwiatkowski et~al.(2019)Kwiatkowski, Palomaki, Redfield, Collins, Parikh, Alberti, Epstein, Polosukhin, Devlin, Lee, Toutanova, Jones, Kelcey, Chang, Dai, Uszkoreit, Le, and Petrov}]{kwiatkowski2019naturalqa}
Tom Kwiatkowski, Jennimaria Palomaki, Olivia Redfield, Michael Collins, Ankur Parikh, Chris Alberti, Danielle Epstein, Illia Polosukhin, Jacob Devlin, Kenton Lee, Kristina Toutanova, Llion Jones, Matthew Kelcey, Ming-Wei Chang, Andrew~M. Dai, Jakob Uszkoreit, Quoc Le, and Slav Petrov. 2019.
\newblock \href {https://doi.org/10.1162/tacl_a_00276} {Natural questions: A benchmark for question answering research}.
\newblock \emph{Transactions of the Association for Computational Linguistics}, 7:452--466.

\bibitem[{Li et~al.(2023)Li, Cheng, Zhao, Nie, and Wen}]{li-etal-2023-halueval}
Junyi Li, Xiaoxue Cheng, Xin Zhao, Jian-Yun Nie, and Ji-Rong Wen. 2023.
\newblock \href {https://doi.org/10.18653/v1/2023.emnlp-main.397} {{H}alu{E}val: A large-scale hallucination evaluation benchmark for large language models}.
\newblock In \emph{Proceedings of the 2023 Conference on Empirical Methods in Natural Language Processing}, pages 6449--6464, Singapore. Association for Computational Linguistics.

\bibitem[{Li et~al.(2024)Li, Wang, Feng, Zhu, Wang, and Chua}]{li2024think}
Moxin Li, Wenjie Wang, Fuli Feng, Fengbin Zhu, Qifan Wang, and Tat-Seng Chua. 2024.
\newblock \href {http://arxiv.org/abs/2403.09972} {Think twice before trusting: Self-detection for large language models through comprehensive answer reflection}.

\bibitem[{Lin et~al.(2024)Lin, Trivedi, and Sun}]{lin2023generating}
Zhen Lin, Shubhendu Trivedi, and Jimeng Sun. 2024.
\newblock \href {https://openreview.net/forum?id=DWkJCSxKU5} {Generating with confidence: Uncertainty quantification for black-box large language models}.
\newblock \emph{Transactions on Machine Learning Research}.

\bibitem[{Liu et~al.(2019)Liu, Ott, Goyal, Du, Joshi, Chen, Levy, Lewis, Zettlemoyer, and Stoyanov}]{liu2019roberta}
Yinhan Liu, Myle Ott, Naman Goyal, Jingfei Du, Mandar Joshi, Danqi Chen, Omer Levy, Mike Lewis, Luke Zettlemoyer, and Veselin Stoyanov. 2019.
\newblock \href {http://arxiv.org/abs/1907.11692} {Roberta: A robustly optimized bert pretraining approach}.

\bibitem[{Lu et~al.(2022)Lu, Zeng, Zhang, Wu, and Li}]{lu-etal-2022-learning}
Yu~Lu, Jiali Zeng, Jiajun Zhang, Shuangzhi Wu, and Mu~Li. 2022.
\newblock \href {https://doi.org/10.18653/v1/2022.acl-long.167} {Learning confidence for transformer-based neural machine translation}.
\newblock In \emph{Proceedings of the 60th Annual Meeting of the Association for Computational Linguistics (Volume 1: Long Papers)}, pages 2353--2364, Dublin, Ireland. Association for Computational Linguistics.

\bibitem[{Luo et~al.(2023)Luo, Xiao, and Ma}]{luo2023zeroresource}
Junyu Luo, Cao Xiao, and Fenglong Ma. 2023.
\newblock \href {http://arxiv.org/abs/2309.02654} {Zero-resource hallucination prevention for large language models}.

\bibitem[{Lyu et~al.(2024)Lyu, Shridhar, Malaviya, Zhang, Elazar, Tandon, Apidianaki, Sachan, and Callison-Burch}]{lyu2024calibrating}
Qing Lyu, Kumar Shridhar, Chaitanya Malaviya, Li~Zhang, Yanai Elazar, Niket Tandon, Marianna Apidianaki, Mrinmaya Sachan, and Chris Callison-Burch. 2024.
\newblock Calibrating large language models with sample consistency.
\newblock \emph{arXiv preprint arXiv:2402.13904}.

\bibitem[{Malinin and Gales(2021)}]{malinin2021uncertainty}
Andrey Malinin and Mark Gales. 2021.
\newblock \href {https://openreview.net/forum?id=jN5y-zb5Q7m} {Uncertainty estimation in autoregressive structured prediction}.
\newblock In \emph{International Conference on Learning Representations}.

\bibitem[{Manakul et~al.(2023)Manakul, Liusie, and Gales}]{manakul2023selfcheckgpt}
Potsawee Manakul, Adian Liusie, and Mark Gales. 2023.
\newblock \href {https://doi.org/10.18653/v1/2023.emnlp-main.557} {{S}elf{C}heck{GPT}: Zero-resource black-box hallucination detection for generative large language models}.
\newblock In \emph{Proceedings of the 2023 Conference on Empirical Methods in Natural Language Processing}, pages 9004--9017, Singapore. Association for Computational Linguistics.

\bibitem[{Mesnard et~al.(2024)Mesnard, Hardin, Dadashi, Bhupatiraju, Pathak, Sifre, Rivière, Kale, Love, Tafti, Hussenot, Sessa, Chowdhery, Roberts, Barua, Botev, Castro-Ros, Slone, Héliou, Tacchetti, Bulanova, Paterson, Tsai, Shahriari, Lan, Choquette-Choo, Crepy, Cer, Ippolito, Reid, Buchatskaya, Ni, Noland, Yan, Tucker, Muraru, Rozhdestvenskiy, Michalewski, Tenney, Grishchenko, Austin, Keeling, Labanowski, Lespiau, Stanway, Brennan, Chen, Ferret, Chiu, Mao-Jones, Lee, Yu, Millican, Sjoesund, Lee, Dixon, Reid, Mikuła, Wirth, Sharman, Chinaev, Thain, Bachem, Chang, Wahltinez, Bailey, Michel, Yotov, Chaabouni, Comanescu, Jana, Anil, McIlroy, Liu, Mullins, Smith, Borgeaud, Girgin, Douglas, Pandya, Shakeri, De, Klimenko, Hennigan, Feinberg, Stokowiec, hui Chen, Ahmed, Gong, Warkentin, Peran, Giang, Farabet, Vinyals, Dean, Kavukcuoglu, Hassabis, Ghahramani, Eck, Barral, Pereira, Collins, Joulin, Fiedel, Senter, Andreev, and Kenealy}]{gemmateam2024gemma}
Thomas Mesnard, Cassidy Hardin, Robert Dadashi, Surya Bhupatiraju, Shreya Pathak, Laurent Sifre, Morgane Rivière, Mihir~Sanjay Kale, Juliette Love, Pouya Tafti, Léonard Hussenot, Pier~Giuseppe Sessa, Aakanksha Chowdhery, Adam Roberts, Aditya Barua, Alex Botev, Alex Castro-Ros, Ambrose Slone, Amélie Héliou, Andrea Tacchetti, Anna Bulanova, Antonia Paterson, Beth Tsai, Bobak Shahriari, Charline~Le Lan, Christopher~A. Choquette-Choo, Clément Crepy, Daniel Cer, Daphne Ippolito, David Reid, Elena Buchatskaya, Eric Ni, Eric Noland, Geng Yan, George Tucker, George-Christian Muraru, Grigory Rozhdestvenskiy, Henryk Michalewski, Ian Tenney, Ivan Grishchenko, Jacob Austin, James Keeling, Jane Labanowski, Jean-Baptiste Lespiau, Jeff Stanway, Jenny Brennan, Jeremy Chen, Johan Ferret, Justin Chiu, Justin Mao-Jones, Katherine Lee, Kathy Yu, Katie Millican, Lars~Lowe Sjoesund, Lisa Lee, Lucas Dixon, Machel Reid, Maciej Mikuła, Mateo Wirth, Michael Sharman, Nikolai Chinaev, Nithum Thain, Olivier Bachem, Oscar Chang,
  Oscar Wahltinez, Paige Bailey, Paul Michel, Petko Yotov, Rahma Chaabouni, Ramona Comanescu, Reena Jana, Rohan Anil, Ross McIlroy, Ruibo Liu, Ryan Mullins, Samuel~L Smith, Sebastian Borgeaud, Sertan Girgin, Sholto Douglas, Shree Pandya, Siamak Shakeri, Soham De, Ted Klimenko, Tom Hennigan, Vlad Feinberg, Wojciech Stokowiec, Yu~hui Chen, Zafarali Ahmed, Zhitao Gong, Tris Warkentin, Ludovic Peran, Minh Giang, Clément Farabet, Oriol Vinyals, Jeff Dean, Koray Kavukcuoglu, Demis Hassabis, Zoubin Ghahramani, Douglas Eck, Joelle Barral, Fernando Pereira, Eli Collins, Armand Joulin, Noah Fiedel, Evan Senter, Alek Andreev, and Kathleen Kenealy. 2024.
\newblock \href {http://arxiv.org/abs/2403.08295} {Gemma: Open models based on gemini research and technology}.

\bibitem[{O{\u{g}}uz et~al.(2024)O{\u{g}}uz, Ciftci, and Bakman}]{oguz-etal-2024-llms}
Metehan O{\u{g}}uz, Yusuf Ciftci, and Yavuz~Faruk Bakman. 2024.
\newblock \href {https://aclanthology.org/2024.sigturk-1.5} {Do {LLM}s recognize me, when {I} is not me: Assessment of {LLM}s understanding of {T}urkish indexical pronouns in indexical shift contexts}.
\newblock In \emph{Proceedings of the First Workshop on Natural Language Processing for Turkic Languages (SIGTURK 2024)}, pages 53--61, Bangkok, Thailand and Online. Association for Computational Linguistics.

\bibitem[{OpenAI(2023)}]{openai2023gpt4}
OpenAI. 2023.
\newblock \href {http://arxiv.org/abs/2303.08774} {{GPT-4 Technical Report}}.

\bibitem[{Ravi et~al.(2024)Ravi, Mielczarek, Kannappan, Kiela, and Qian}]{ravi2024lynxopensourcehallucination}
Selvan~Sunitha Ravi, Bartosz Mielczarek, Anand Kannappan, Douwe Kiela, and Rebecca Qian. 2024.
\newblock \href {http://arxiv.org/abs/2407.08488} {Lynx: An open source hallucination evaluation model}.

\bibitem[{Touvron et~al.(2023)Touvron, Martin, Stone, Albert, Almahairi, Babaei, Bashlykov, Batra, Bhargava, Bhosale, Bikel, Blecher, Ferrer, Chen, Cucurull, Esiobu, Fernandes, Fu, Fu, Fuller, Gao, Goswami, Goyal, Hartshorn, Hosseini, Hou, Inan, Kardas, Kerkez, Khabsa, Kloumann, Korenev, Koura, Lachaux, Lavril, Lee, Liskovich, Lu, Mao, Martinet, Mihaylov, Mishra, Molybog, Nie, Poulton, Reizenstein, Rungta, Saladi, Schelten, Silva, Smith, Subramanian, Tan, Tang, Taylor, Williams, Kuan, Xu, Yan, Zarov, Zhang, Fan, Kambadur, Narang, Rodriguez, Stojnic, Edunov, and Scialom}]{touvron2023llama2}
Hugo Touvron, Louis Martin, Kevin Stone, Peter Albert, Amjad Almahairi, Yasmine Babaei, Nikolay Bashlykov, Soumya Batra, Prajjwal Bhargava, Shruti Bhosale, Dan Bikel, Lukas Blecher, Cristian~Canton Ferrer, Moya Chen, Guillem Cucurull, David Esiobu, Jude Fernandes, Jeremy Fu, Wenyin Fu, Brian Fuller, Cynthia Gao, Vedanuj Goswami, Naman Goyal, Anthony Hartshorn, Saghar Hosseini, Rui Hou, Hakan Inan, Marcin Kardas, Viktor Kerkez, Madian Khabsa, Isabel Kloumann, Artem Korenev, Punit~Singh Koura, Marie-Anne Lachaux, Thibaut Lavril, Jenya Lee, Diana Liskovich, Yinghai Lu, Yuning Mao, Xavier Martinet, Todor Mihaylov, Pushkar Mishra, Igor Molybog, Yixin Nie, Andrew Poulton, Jeremy Reizenstein, Rashi Rungta, Kalyan Saladi, Alan Schelten, Ruan Silva, Eric~Michael Smith, Ranjan Subramanian, Xiaoqing~Ellen Tan, Binh Tang, Ross Taylor, Adina Williams, Jian~Xiang Kuan, Puxin Xu, Zheng Yan, Iliyan Zarov, Yuchen Zhang, Angela Fan, Melanie Kambadur, Sharan Narang, Aurelien Rodriguez, Robert Stojnic, Sergey Edunov, and Thomas
  Scialom. 2023.
\newblock \href {http://arxiv.org/abs/2307.09288} {Llama 2: Open foundation and fine-tuned chat models}.

\bibitem[{Ulmer et~al.(2024)Ulmer, Gubri, Lee, Yun, and Oh}]{ulmer2024calibrating}
Dennis Ulmer, Martin Gubri, Hwaran Lee, Sangdoo Yun, and Seong Oh. 2024.
\newblock \href {https://doi.org/10.18653/v1/2024.acl-long.824} {Calibrating large language models using their generations only}.
\newblock In \emph{Proceedings of the 62nd Annual Meeting of the Association for Computational Linguistics (Volume 1: Long Papers)}, pages 15440--15459, Bangkok, Thailand. Association for Computational Linguistics.

\bibitem[{van~den Oord et~al.(2016)van~den Oord, Kalchbrenner, Espeholt, kavukcuoglu, Vinyals, and Graves}]{NIPS2016_b1301141}
Aaron van~den Oord, Nal Kalchbrenner, Lasse Espeholt, koray kavukcuoglu, Oriol Vinyals, and Alex Graves. 2016.
\newblock \href {https://proceedings.neurips.cc/paper_files/paper/2016/file/b1301141feffabac455e1f90a7de2054-Paper.pdf} {Conditional image generation with pixelcnn decoders}.
\newblock In \emph{Advances in Neural Information Processing Systems}, volume~29. Curran Associates, Inc.

\bibitem[{Vaswani(2017)}]{vaswani2017attention}
A~Vaswani. 2017.
\newblock Attention is all you need.
\newblock \emph{Advances in Neural Information Processing Systems}.

\bibitem[{Wang et~al.(2024)Wang, Reddy, Mujahid, Arora, Rubashevskii, Geng, Afzal, Pan, Borenstein, Pillai, Augenstein, Gurevych, and Nakov}]{wang2024factcheckbench}
Yuxia Wang, Revanth~Gangi Reddy, Zain~Muhammad Mujahid, Arnav Arora, Aleksandr Rubashevskii, Jiahui Geng, Osama~Mohammed Afzal, Liangming Pan, Nadav Borenstein, Aditya Pillai, Isabelle Augenstein, Iryna Gurevych, and Preslav Nakov. 2024.
\newblock \href {http://arxiv.org/abs/2311.09000} {Factcheck-bench: Fine-grained evaluation benchmark for automatic fact-checkers}.

\bibitem[{Ye et~al.(2023)Ye, Chen, Xu, Zu, Shao, Liu, Cui, Zhou, Gong, Shen, Zhou, Chen, Gui, Zhang, and Huang}]{ye2023comprehensive}
Junjie Ye, Xuanting Chen, Nuo Xu, Can Zu, Zekai Shao, Shichun Liu, Yuhan Cui, Zeyang Zhou, Chao Gong, Yang Shen, Jie Zhou, Siming Chen, Tao Gui, Qi~Zhang, and Xuanjing Huang. 2023.
\newblock \href {http://arxiv.org/abs/2303.10420} {A comprehensive capability analysis of {GPT-3 and GPT-3.5} series models}.

\bibitem[{Zhang et~al.(2023)Zhang, Li, Das, Malin, and Kumar}]{zhang2024sac3}
Jiaxin Zhang, Zhuohang Li, Kamalika Das, Bradley Malin, and Sricharan Kumar. 2023.
\newblock \href {https://doi.org/10.18653/v1/2023.findings-emnlp.1032} {{SAC}$^3$: Reliable hallucination detection in black-box language models via semantic-aware cross-check consistency}.
\newblock In \emph{Findings of the Association for Computational Linguistics: EMNLP 2023}, pages 15445--15458, Singapore. Association for Computational Linguistics.

\bibitem[{Zhao et~al.(2024)Zhao, Yan, Sun, Xing, Meng, Wang, Cheng, Ren, and Yin}]{zhao2023knowing}
Yukun Zhao, Lingyong Yan, Weiwei Sun, Guoliang Xing, Chong Meng, Shuaiqiang Wang, Zhicong Cheng, Zhaochun Ren, and Dawei Yin. 2024.
\newblock \href {https://doi.org/10.18653/v1/2024.naacl-long.390} {Knowing what {LLM}s {DO} {NOT} know: A simple yet effective self-detection method}.
\newblock In \emph{Proceedings of the 2024 Conference of the North American Chapter of the Association for Computational Linguistics: Human Language Technologies (Volume 1: Long Papers)}, pages 7051--7063, Mexico City, Mexico. Association for Computational Linguistics.

\end{thebibliography}
\bibliographystyle{acl_natbib}

\appendix
\clearpage

\section{Details of non-English Languages Experiment}

\textbf{Preparing calibration data for LARS:} We translate the same 13k question-ground truth pairs from the train split of TriviaQA to Turkish, German, and Spanish using the Googletrans library\footnote{https://py-googletrans.readthedocs.io}. Then, we apply the same procedure as we do for English: Make the LLM generate 6 answers to the question, ensuring
the most likely generation is included. The labels for each QA pair are obtained by using GPT-3.5-turbo. To train LARS, we utilize unique question-response pairs. 

\noindent\textbf{Preparing test data:} To test the performance of varying scoring functions in these languages, we translate the question-ground truth pairs of test samples of TriviaQA. After having the translated test set, the Entropy UE metric is calculated by using various scoring functions. 

\noindent\textbf{Prompts for the LLM:} The prompts for the LLM to generate the answers are also translated into the corresponding languages to make sure it provides answers in the target language. Llama3-8b is used for this experiment since it is known to be trained in these languages. Prompts are provided below.

To generate answers in Turkish:
\begin{verbatim}
System: Sen yardımcı, saygılı ve dürüst 
bir asistansın. Sorularımı Türkçe olacak 
şekilde net, kısa ve öz cevapla.
User: {question}\end{verbatim}

To generate answers in German:
\begin{verbatim}
System: Du bist ein hilfreicher Assistent. 
Geben Sie auf die gestellten Fragen präzise, 
kurze Antworten in einem Satz auf Deutsch.
User: {question}\end{verbatim}

To generate answers in Spanish:
\begin{verbatim}
System: Eres un asistente servicial, 
respetuoso y honesto. Das respuestas 
precisas, breves y de una sola oración a 
las preguntas que se te dan en español.
User: {question}\end{verbatim}

The English translation of the above prompts is as follows:
\begin{verbatim}
System: You are a helpful, respectful
and honest assistant. Give short and 
precise answers to given 
questions in {target_language}.
User: {question}\end{verbatim}

\noindent\textbf{Prompt for GPT-3.5-turbo:} The following prompt is used for GPT-3.5-turbo to obtain labels:
\begin{verbatim}
You will behave as a question answer 
evaluator. I will give you a question, 
the ground truth of the question, and 
a generated answer by a language model 
in {target_language}. You will output 
"correct" if the generated answer is 
correct regarding question and ground 
truth. Otherwise, output "false".
Question: {question}, 
Ground Truth: {gt_answer},
Generated Answer: {generation}
\end{verbatim}

\section{Details of LARS training}

We use the pre-trained RoBERTa-base model with a single logit fully-connected layer added to the end. Binary cross entropy loss is used, while the optimizer is AdamW with a learning rate of $5e-6$. The model is trained for 5 epochs. We did a search for batch size in the set of $\{4, 8, 16, 32\}$ and found the optimal batch size as 8 and used it in all of the experiments. The search set for learning rate was $\{1e-6, 5e-6, 1e-5, 5e-4, 1e-4, 5e-4\}$. Lastly, we explored training the model for more epochs (up to 10); however, after epoch 5, we observed overfitting.

The embedding vectors of probability tokens are initialized as few-hot as explained in Section~\ref{method} and kept frozen during the training of the model. We also experimented with training those vectors as well as initializing them as fully non-zero random vectors. We observed that the mentioned few-hot strategy gives superior and more stable results. On the other hand, for the additive probability association approach explained in Section~\ref{sec:v1-v2}, initializing the embedding vectors as few-hot while keeping them trainable gave the best performance.  

\section{Additional Experiments}\label{additional_experiments}

\begin{table*}[!htbp]
\centering
\fontsize{7.5}{8}\selectfont
\begin{tabular}{l|c|l| cc | cc |cc| cc|cc }
\toprule
  &UE Method & Scoring & \multicolumn{2}{c}{\textbf{Llama2-7b}} & \multicolumn{2}{c}{\textbf{Llama3-8b}} & \multicolumn{2}{c}{\textbf{Mistral-7b}}& \multicolumn{2}{c}{\textbf{Gemma-7b}} & \multicolumn{2}{c}{\textbf{Llama2-13b}} \\
   && Function & AUROC & PRR & AUROC & PRR & AUROC & PRR & AUROC & PRR & AUROC & PRR \\
\midrule[\heavyrulewidth]
\multirow{24}{*}{\rotatebox{90}{\textbf{TriviaQA}}}
&\textbf{Lex. Sim.}          &-&0.647 &0.374	&0.683 &0.483	&0.720 &0.517	&0.597 &0.227 &0.611 &0.314  \\
&\textbf{\# Sem. Gr.}     &-&0.792 &0.571	&0.819 &0.671	&0.757 &0.521	&0.744 &0.454 &0.776 &0.557\\
&$\mathbf{p}$\textbf{(True)} &-&0.616 &0.267	&0.842 &0.733    &0.805 &0.653	&0.517 &0.023 &0.650&0.392 \\
&\textbf{SAPLMA}             &-&0.741 &0.484 &0.736 &0.541    &0.785 &0.614	&0.658 &0.373  & 0.757& 0.594\\
&\textbf{Eccentricity}           &-&0.812 &0.629 &0.853 &0.756    &0.818 &0.664	&0.764 &0.496 &0.813 &0.633\\
&\textbf{Degree Matrix}          &-&0.812 &0.620 &0.851 &0.746    &0.820 &0.658	&0.766 &0.511 &0.817 &0.646\\
  \cmidrule{2-13}
&\multirow{4}{*}{\textbf{Confidence}}
&LNS         &0.697 & 0.481	&0.748 &0.600	&0.722 &0.533	&0.628 &0.281  &0.655 &0.389\\
&&MARS       &0.751 &0.576	&0.799 &0.676	&0.745 &0.593	&0.638 &0.305  &0.641 &0.381 \\
&&TokenSAR   &0.747 &0.572	&0.792 &0.674	&0.747 &0.584	&0.688 &0.386  &0.728 &0.527 \\
&&LARS       &\textbf{0.851} &\textbf{0.760}	&\textbf{0.872} &\textbf{0.817}	&\textbf{0.844} &\textbf{0.759}	&\textbf{0.819} &\textbf{0.690}  &\textbf{0.846} &\textbf{0.766} \\
  \cmidrule{2-13}
&\multirow{4}{*}{\textbf{Entropy}}
&LNS         &0.692 &0.461	&0.747 &0.594	&0.738 &0.563	&0.633 &0.286  &0.669 &0.404 \\
&&MARS       &0.736 &0.547	&0.801 &0.672	&0.755 &0.602	&0.659 &0.336  &0.672 &0.421\\
&&TokenSAR   &0.734 &0.546	&0.793 &0.676	&0.763 &0.610	&0.694 &0.398  &0.733 &0.528 \\
&&LARS       &\textbf{0.842} &\textbf{0.748}	&\textbf{0.864} &\textbf{0.804}	&\textbf{0.849} &\textbf{0.773}	&\textbf{0.818} &\textbf{0.690}  &\textbf{0.853} &\textbf{0.779} \\
  \cmidrule{2-13}
&\multirow{4}{*}{\textbf{SE}}
&LNS         &0.795 &0.627	&0.835 &0.733	&0.810 &0.670	&0.749 &0.475  &0.800 &0.617 \\
&&MARS       &0.797 &0.645	&0.835 &0.742	&0.810 &0.681	&0.749 &0.482  &0.794 &0.615 \\
&&TokenSAR   &0.796 &0.640	&0.839 &0.747	&0.813 &0.681	&0.753 &0.493  &0.806 &0.639 \\
&&LARS       &\textbf{0.849} &\textbf{0.745}	&\textbf{0.866} &\textbf{0.811}	&\textbf{0.854} &\textbf{0.782}	&\textbf{0.821} &\textbf{0.699}  &\textbf{0.866} &\textbf{0.797} \\
\cmidrule{2-13}
&\multirow{4}{*}{\textbf{SentSAR}}
&LNS         &0.784 &0.611	&0.825 &0.723   &0.796 &0.652	&0.728 &0.448  &0.778 &0.593 \\
&&MARS       &0.794 &0.636	&0.838 &0.746	&0.802 &0.668	&0.731 &0.456  &0.773 &0.590 \\
&&TokenSAR   &0.790 &0.633	&0.840 &0.750	&0.805 &0.669	&0.741 &0.475  &0.791 &0.618\\
&&LARS       &\textbf{0.850} &\textbf{0.763}	&\textbf{0.879} &\textbf{0.823}	&\textbf{0.855} &\textbf{0.773}	&\textbf{0.823} &\textbf{0.685}  &\textbf{0.859} &\textbf{0.770}\\
\midrule[\heavyrulewidth]
\multirow{24}{*}{\rotatebox{90}{\textbf{NaturalQA}}}
& \textbf{Lex. Sim.}   &-&0.600 &0.263	&0.651 &0.373	&0.637 &0.340	&0.585 &0.163 &0.604 &0.261\\
&\textbf{\# Sem. Gr.}   &-&0.705 &0.379	&0.736 &0.448	&0.675 &0.283	&0.686 &0.276 &0.709 &0.377\\
&$\mathbf{p}$\textbf{(True)}            &-&0.561 &0.90	&0.761 &0.561  &0.727 &0.509	&0.647 &0.247  &0.562 &0.131 \\
&\textbf{SAPLMA}          &-&0.691 &0.397	&0.713 &0.443 &0.723 &0.458	&0.657 &0.289  & 0.594&0.410\\
&\textbf{Eccentricity}             &-&0.727 &0.431 &0.775 &0.567    &0.727 &0.480	&0.713 &0.368 &0.741 &0.482\\
&\textbf{Degree Matrix}             &-&0.727 &0.435 &0.771 &0.554    &0.732 &0.483	&0.715 &0.358 &0.742 &0.487\\
  \cmidrule{2-13}
&\multirow{4}{*}{\textbf{Confidence}}
&LNS         &0.677 &0.384	&0.697 &0.449	&0.666 &0.390	&0.610 &0.189  &0.648 &0.338 \\
&&MARS       &0.699 &0.411	&0.717 &0.477	&0.678 &0.407	&0.615 &0.198  &0.631 &0.311 \\
&&TokenSAR   &0.703 &0.431	&0.717 &0.476	&0.682 &0.426	&0.643 &0.249  &0.677 &0.393 \\
&&LARS       &\textbf{0.780} &\textbf{0.581}	&\textbf{0.812} &\textbf{0.654}	&\textbf{0.782} &\textbf{0.599}	&\textbf{0.794} &\textbf{0.541}  &\textbf{0.772} &\textbf{0.574} \\
\cmidrule{2-13}
&\multirow{4}{*}{\textbf{Entropy}}
&LNS         &0.661 &0.559	&0.698 &0.449	&0.679 &0.419	&0.611 &0.202  &0.656 &0.355 \\
&&MARS       &0.681 &0.379	&0.707 &0.475	&0.691 &0.447	&0.616 &0.199  &0.636 &0.304 \\
&&TokenSAR   &0.683 &0.392	&0.714 &0.477	&0.694 &0.451	&0.644 &0.261  &0.686 &0.410 \\
&&LARS       &\textbf{0.775} &\textbf{0.573}	&\textbf{0.805} &\textbf{0.652}	&\textbf{0.781} &\textbf{0.595}	&\textbf{0.785} &\textbf{0.529}  &\textbf{0.773} &\textbf{0.574} \\
\cmidrule{2-13}
&\multirow{4}{*}{\textbf{SE}}
&LNS         &0.721 &0.432	&0.759 &0.548	&0.727 &0.499	&0.700 &0.332  &0.733 &0.471 \\
&&MARS       &0.720 &0.440	&0.750 &0.546	&0.725 &0.493	&0.705 &0.336  &0.723 &0.440 \\
&&TokenSAR   &0.721 &0.443	&0.756 &0.544	&0.726 &0.498	&0.700 &0.340  &0.736 &0.485 \\
&&LARS       &\textbf{0.772} &\textbf{0.569}	&\textbf{0.794} &\textbf{0.638}	&\textbf{0.778} &\textbf{0.591}	&\textbf{0.785} &\textbf{0.548}  &\textbf{0.779} &\textbf{0.583}\\
\cmidrule{2-13}
&\multirow{4}{*}{\textbf{SentSAR}}
&LNS         &0.712 &0.423	&0.752 &0.543    &0.721 &0.487	&0.680 &0.297  &0.725 &0.468 \\
&&MARS       &0.718 &0.435	&0.752 &0.550	&0.722 &0.492	&0.689 &0.301  &0.714 &0.443 \\
&&TokenSAR   &0.718 &0.438	&0.756 &0.551	&0.727 &0.496	&0.684 &0.309  &0.732 &0.485\\
&&LARS       &\textbf{0.779} &\textbf{0.579}	&\textbf{0.814} &\textbf{0.665}	&\textbf{0.789} &\textbf{0.616}	&\textbf{0.793} &\textbf{0.551}  &\textbf{0.784} &\textbf{0.583}\\

\bottomrule
\end{tabular}
\caption{AUROC and PRR scores of UE methods on TriviaQA and NaturalQA.}
\vskip -0.15in
\label{tab:main_results_big}
\end{table*}
\begin{table*}[!htbp]
\centering
\fontsize{7.5}{8}\selectfont
\begin{tabular}{c|l| cc |cc|cc|cc|cc }
\toprule
  UE Method & Scoring & \multicolumn{2}{c}{\textbf{Llama2-7b}} & \multicolumn{2}{c}{\textbf{Llama3-8b}} & \multicolumn{2}{c}{\textbf{Mistral-7b}}& \multicolumn{2}{c}{\textbf{Gemma-7b}} & \multicolumn{2}{c}{\textbf{Llama2-13b}} \\
   & Function & AUROC & PRR & AUROC & PRR & AUROC & PRR & AUROC & PRR & AUROC & PRR \\
\midrule[\heavyrulewidth]

\textbf{Lex. Sim.}   & -  &0.643 &0.310	&0.640 &0.321	&0.645 &0.312	&0.608 &0.214 &0.624  &0.261	  \\
\textbf{\# Sem. Gr.} & -  &0.612 &0.138	&0.599 &0.143	&0.601 &0.184	&0.630 &0.213 &0.587  &0.157		  \\
$\mathbf{p}$\textbf{(True)}  &-&0.558 &0.078	&0.636 &0.290	  &0.667 &0.358	  &0.552 &0.041	&0.580 &0.171	\\
\textbf{Eccentricity} & -      &0.680 &0.375 &0.674 &0.386	&0.662 &0.333	&0.606 &0.203	&0.686  &0.358	  \\
\textbf{Degree Matrix} & -  &0.683 &0.380 &0.676 &0.384	&0.662 &0.326	&0.611 &0.195	&0.682  &0.364	  \\

\midrule[\heavyrulewidth]
\multirow{6}{*}{\textbf{Confidence}}
&LNS        &0.656 &0.329	&0.645 &0.324		&0.634 &0.305		&0.625 &0.246	&0.602  &0.233	 \\
&MARS       &0.669 &0.349	&0.649 &0.333		&0.637 &0.316		&0.627 &0.258	&0.585  &0.199	 \\
&TokenSAR   &0.664 &0.345	&0.656 &0.347		&0.640 &0.320		&0.657 &0.287	&0.615  &0.248	  \\
&LARS (T)   &0.701 &0.413   &0.704 &0.423		&0.681 &0.399		&0.710 &0.422	&0.675 &0.368	  \\
&LARS (N)   &0.701 & 0.385		&0.690 &	0.413	&0.682 &0.358		&\textbf{0.732} &0.439	&\textbf{0.683}  &\textbf{0.384}	  \\
&LARS (T+N) &\textbf{0.715} &\textbf{0.430}		&\textbf{0.713} &\textbf{0.464}		&\textbf{0.686} &\textbf{0.406}		&0.726 &\textbf{0.442}	&0.676  &0.367	  \\
\midrule[\heavyrulewidth]
\multirow{6}{*}{\textbf{Entropy}}
&LNS        &0.656 &0.332		&0.650 &0.340	&0.647 &0.323	&0.638 &0.272	&0.625  &0.259	 \\
&MARS       &0.675 &0.354		&0.657 &0.349	&0.647 &0.328	&0.656 &0.293	&0.602  &0.219	 \\
&TokenSAR   &0.668 &0.351		&0.661 &0.355	&0.649 &0.330	&0.665 &0.307	&0.630  &0.267	  \\
&LARS (T)   &0.705 &0.433	&\textbf{0.705} &0.430	&0.686 &0.405	&0.710 &0.428	&0.681  &0.388	  \\
&LARS (N)   &0.712 &0.418		&0.690 &0.428		&0.691 &0.393		&\textbf{0.731} &\textbf{0.438}	 &\textbf{0.687}  &\textbf{0.401}	 \\
&LARS (T+N) &\textbf{0.714} &\textbf{0.441}		&0.703 &\textbf{0.456}		&\textbf{0.693} &\textbf{0.422}		&0.717 &0.430	&0.677  &0.376	  \\
\midrule[\heavyrulewidth]
\multirow{6}{*}{\textbf{SE}}
&LNS        &0.672 &0.360		&0.664 &0.366	&0.665 &0.353		&0.675 &0.334	&0.644  &0.297	 \\
&MARS       &0.679 &0.367		&0.667 &0.370	&0.665 &0.354		&0.679 &0.340	 &0.632  &0.267	 \\
&TokenSAR   &0.674 &0.365		&0.667 &0.372	&0.663 &0.351		&0.680 &0.343	 &0.647  &0.298	 \\
&LARS (T)   &0.707 &0.439		&\textbf{0.697} &0.428 &0.686 &0.407		&0.710 &0.431	&0.680  &0.388	 \\
&LARS (N)   &0.709 &0.422		&0.685 &0.426		&0.693 &0.402		&\textbf{0.726} &0.437	&\textbf{0.684}  &\textbf{0.400}	  \\
&LARS (T+N) &\textbf{0.711} &\textbf{0.440}		&0.694 &\textbf{0.449}		&\textbf{0.697} &\textbf{0.430}		&0.719 &\textbf{0.440}	&0.678  &0.382	 \\
\midrule[\heavyrulewidth]
\multirow{6}{*}{\textbf{SentSAR}}
&LNS        &0.703 &0.406		&0.687 &0.400	&0.677 &0.362	&0.691 &0.356	&0.672  &0.336	\\
&MARS       &0.705 &0.408		&0.692 &0.406	&0.677 &0.365	&0.700 &0.365	 &0.662  &0.313	 \\
&TokenSAR   &0.704 &0.407		&0.691 &0.406	&0.678 &0.363	&0.698 &0.364	&0.671  &0.333	 \\
&LARS (T)   &0.714 &0.445		&0.718 &0.455	&0.693 &0.408	&0.724 &0.457	&0.695  &0.403	 \\
&LARS (N)   &\textbf{0.730} &0.465		&0.705 &0.455		&\textbf{0.705}	 &\textbf{0.423}	&\textbf{0.747} &\textbf{0.484}	&\textbf{0.701}  &\textbf{0.421}	  \\
&LARS (T+N) &0.728 &\textbf{0.471}	 &\textbf{0.721} &\textbf{0.484}		&{0.698} &0.409		&0.732 &0.467 &0.692  &0.404		\\
\bottomrule
\end{tabular}
\caption{AUROC and PRR performance of UE methods on WebQA dataset. LARS models are trained with TriviaQA~(T) and/or NaturalQA (N).}
\vskip -0.2in
\label{tab:webqa_big}
\end{table*}
\begin{table*}[!htbp]
\centering
\fontsize{7.5}{8}\selectfont
\begin{tabular}{c|l| cc |cc|cc|cc|cc }
\toprule
  UE Method & Scoring & \multicolumn{2}{c}{\textbf{Llama2-7b}} & \multicolumn{2}{c}{\textbf{Llama3-8b}} & \multicolumn{2}{c}{\textbf{Mistral-7b}}& \multicolumn{2}{c}{\textbf{Gemma-7b}} & \multicolumn{2}{c}{\textbf{Llama2-13b}} \\
   & Function & AUROC & PRR & AUROC & PRR & AUROC & PRR & AUROC & PRR & AUROC & PRR \\
\midrule[\heavyrulewidth]

\textbf{Lex. Sim.}	&-	&0.444	&0	&0.632	&0.272	&0.537	&0.019	&0.544	&0.080	&0.551	&0.110	\\
\textbf{\# Sem. Gr.}	&-	&0.513	&0	&0.584	&0.138	&0.532	&0.037	&0.566	&0.114	&0.561	&0.065	\\
$\mathbf{p}$\textbf{(True)}	&-	&0.540	&0.099	&0.797	&0.623	&0.665	&0.238	&0.486	&0	&0.501	&0	\\
\textbf{Eccentricity}	&-	&0.547	&0.049	&0.664	&0.384	&0.584	&0.109	&0.595	&0.146	&0.600	&0.163	\\
\textbf{Degree Matrix}	&-	&0.535	&0.056	&0.667	&0.667	&0.604	&0.165	&0.584	&0.117	&0.605	&0.179	\\

\midrule[\heavyrulewidth]
\multirow{7}{*}{\textbf{Confidence}}
&LNS	&0.570	&0.031	&0.686	&0.390	&0.567	&0.072	&0.556	&0.370	&0.615	&0.196	\\
	&MARS	&0.567	&0.010	&0.713	&0.438	&0.568	&0.076	&0.541	&0.099	&0.562	&0.114	\\
	&TokenSAR	&0.579	&0.045	&0.719	&0.460	&0.619	&0.156	&0.579	&0.161	&0.636	&0.233	\\
	&LARS(G)	&\textbf{0.720}	&\textbf{0.319}	&\textbf{0.836}	&\textbf{0.711}	&\textbf{0.708}	&\textbf{0.350}	&\textbf{0.706}	&\textbf{0.370}	&\textbf{0.738}	&\textbf{0.497}	\\
	&LARS(T)	&0.582	&0.091	&0.683	&0.379	&0.637	&0.197	&0.554	&0.081	&0.584	&0.137	\\
	&LARS(N)	&0.578	&0.087	&0.695	&0.391	&0.603	&0.149	&0.600	&0.144	&0.641	&0.239	\\
	&LARS(T+N)	&0.603	&0.097	&0.684	&0.348	&0.630	&0.188	&0.676	&0.114	&0.635	&0.218	\\
\midrule[\heavyrulewidth]
\multirow{7}{*}{\textbf{Entropy}}
&LNS	&0.511	&0	&0.643	&0.308	&0.571	&0.090	&0.570	&0.124	&0.574	&0.123	\\
&MARS	&0.509	&0	&0.668	&0.367	&0.573	&0.088	&0.559	&0.103	&0.562	&0.086	\\
&TokenSAR	&0.537	&0	&0.665	&0.369	&0.618	&0.148	&0.577	&0.131	&0.597	&0.156	\\
&LARS(G)	&\textbf{0.701}	&\textbf{0.300}	&\textbf{0.759}	&\textbf{0.579}	&\textbf{0.684}	&\textbf{0.316}	&\textbf{0.681}	&\textbf{0.328}	&\textbf{0.706}	&\textbf{0.408}	\\
&LARS(T)	&0.579	&0.082	&0.641	&0.330	&0.624	&0.194	&0.555	&0.070	&0.586	&0.172	\\
&LARS(N)	&0.542	&0.047	&0.646	&0.319	&0.623	&0.166	&0.557	&0.089	&0.594	&0.182	\\
&LARS(T+N)	&0.586	&0.083	&0.632	&0.291	&0.624	&0.182	&0.562	&0.081	&0.604	&0.191	\\
\midrule[\heavyrulewidth]
\multirow{7}{*}{\textbf{SE}}
&LNS	&0.516	&0	&0.633	&0.321	&0.560	&0.076	&0.588	&0.141	&0.587	&0.153	\\
&MARS	&0.513	&0	&0.640	&0.344	&0.563	&0.080	&0.586	&0.134	&0.583	&0.122	\\
&TokenSAR	&0.526	&0.005	&0.638	&0.344	&0.578	&0.102	&0.588	&0.148	&0.592	&0.171	\\
&LARS(G)	&\textbf{0.675}	&\textbf{0.267}	&\textbf{0.715}	&\textbf{0.528}	&\textbf{0.663}	&\textbf{0.310}	&\textbf{0.679}	&\textbf{0.345}	&\textbf{0.697}	&\textbf{0.383}	\\
&LARS(T)	&0.565	&0.068	&0.639	&0.325	&0.598	&0.173	&0.573	&0.096	&0.593	&0.186	\\
&LARS(N)	&0.537	&0.042	&0.635	&0.319	&0.598	&0.141	&0.581	&0.128	&0.598	&0.194	\\
&LARS(T+N)	&0.572	&0.072	&0.633	&0.298	&0.605	&0.170	&0.579	&0.112	&0.608	&0.209	\\
\midrule[\heavyrulewidth]
\multirow{7}{*}{\textbf{SentSAR}}
&LNS	&0.559	&0.007	&0.698	&0.440	&0.622	&0.167	&0.580	&0.335	&0.634	&0.220	\\
&MARS	&0.545	&0	&0.712	&0.467	&0.618	&0.157	&0.574	&0.128	&0.622	&0.187	\\
&TokenSAR	&0.569	&0.027	&0.712	&0.477	&0.645	&0.189	&0.581	&0.134	&0.640	&0.226	\\
&LARS(G)	&\textbf{0.699}	&\textbf{0.300}	&\textbf{0.772}	&\textbf{0.613}	&\textbf{0.695}	&\textbf{0.338}	&\textbf{0.681}	&\textbf{0.335}	&\textbf{0.712}	&\textbf{0.419}	\\
&LARS(T)	&0.579	&0.080	&0.662	&0.383	&0.639	&0.214	&0.562	&0.081	&0.591	&0.170	\\
&LARS(N)	&0.551	&0.046	&0.671	&0.389	&0.641	&0.209	&0.569	&0.109	&0.605	&0.202	\\
&LARS(T+N)	&0.587	&0.081	&0.645	&0.323	&0.636	&0.207	&0.574	&0.090	&0.612	&0.199	\\
\bottomrule
\end{tabular}
\caption{AUROC and PRR performance of UE methods on GSM8K dataset. LARS models are trained with GSM8K~(G) or TriviaQA~(T) and/or NaturalQA (N).}
\vskip -0.2in
\label{tab:gsm8k}
\end{table*}
\begin{table*}[!htbp]
\centering
\fontsize{10.5}{12}\selectfont
\begin{tabular}{c|l| cccc }
\toprule
 UE Method & Scoring Function & \textbf{Llama2-7b} & \textbf{Llama3-8b} & \textbf{Mistral-7b}& \textbf{Gemma-7b} \\
\midrule[\heavyrulewidth]

\multirow{4}{*}{\textbf{Confidence}}
&Best Score of Baselines     &0.7032	&0.7136	&0.6823	&0.6433   \\
&LARS (N)          &0.7685	&0.7940	&0.7765	&\textbf{0.7919}  \\
&LARS (T)          &0.7455	&0.7689	&0.7365	&0.7415  \\
&LARS (T+N)        &\textbf{0.7731}	&\textbf{0.7997}	&\textbf{0.7774}	&0.7838  \\
\midrule[\heavyrulewidth]
\multirow{4}{*}{\textbf{Entropy}}
&Best Score of Baselines       &0.6831	&0.7144	&0.6944	&0.6439   \\
&LARS (N)          &\textbf{0.7655}	&\textbf{0.7936}	&\textbf{0.7781}	&\textbf{0.7832}  \\
&LARS (T)          &0.7434	&0.7736	&0.7392	&0.7431 \\
&LARS (T+N)        &0.7629	&0.7918	&0.7761	&0.7759  \\
\midrule[\heavyrulewidth]
\multirow{4}{*}{\textbf{SE}}
&Best Score of Baselines       &0.7210	&0.7591	&0.7272	&0.7049  \\
&LARS (N)          &\textbf{0.7665}	&\textbf{0.7873}	&\textbf{0.7770}	&\textbf{0.7845}  \\
&LARS (T)          &0.7511	&0.7750	&0.7497	&0.7594  \\
&LARS (T+N)        &0.7635	&0.7849	&0.7766	&0.7804  \\
\midrule[\heavyrulewidth]
\multirow{4}{*}{\textbf{SentSAR}}
&Best Score of Baselines       &0.7177 &0.7563 &0.7268 &0.6891  \\
&LARS (N)          &0.7709 &\textbf{0.8034} &\textbf{0.7880} &\textbf{0.7900} \\
&LARS (T)          &0.7496 &0.7845 &0.7492 &0.7508  \\
&LARS (T+N)        &\textbf{0.7714} &0.8031 &0.7832 &0.7812  \\
\bottomrule
\end{tabular}
\caption{OOD data experiments on NaturalQA dataset with AUROC score. LARS models are trained with TriviaQA~(T) and/or NaturalQA (N).}
\label{tab:ood_natural}
\end{table*}
\begin{table*}[!htbp]
\centering
\fontsize{10.5}{12}\selectfont
\begin{tabular}{c|l| cccc }
\toprule
 UE Method & Scoring Function & \textbf{Llama2-7b} & \textbf{Llama3-8b} & \textbf{Mistral-7b}& \textbf{Gemma-7b} \\
\midrule[\heavyrulewidth]

\multirow{4}{*}{\textbf{Confidence}}
&Best Score of Baselines       &0.7510	&0.7994	&0.7468	&0.6883  \\
&LARS (T)          &\textbf{0.8505}	&\textbf{0.8721}	&\textbf{0.8443}	&\textbf{0.8191}  \\
&LARS (N)          &0.7780	&0.8243	&0.7893	&0.7678 \\
&LARS (T+N)        &0.8414	&0.8620	&0.8305	&0.8060  \\
\midrule[\heavyrulewidth]
\multirow{4}{*}{\textbf{Entropy}}
&Best Score of Baselines       &0.7356	&0.8012	&0.7634	&0.6941  \\
&LARS (T)          &\textbf{0.8416}	&\textbf{0.8642}	&\textbf{0.8488}	&\textbf{0.8184}  \\
&LARS (N)          &0.7852	&0.8348	&0.8090	&0.7760  \\
&LARS (T+N)        &0.8354	&0.8602	&0.8373	&0.8092  \\
\midrule[\heavyrulewidth]
\multirow{4}{*}{\textbf{SE}}
&Best Score of Baselines       &0.7973	&0.8388	&0.8132	&0.7528 \\
&LARS (T)          &\textbf{0.8488}	&\textbf{0.8662}	&\textbf{0.8541}	&\textbf{0.8214}  \\
&LARS (N)          &0.8181	&0.8515	&0.8349	&0.7968  \\
&LARS (T+N)        &0.8457	&0.8621	&0.8493	&0.8157  \\
\midrule[\heavyrulewidth]
\multirow{4}{*}{\textbf{SentSAR}}
&Best Score of Baselines       &0.7940	&0.8402	&0.8050&0.7411\\
&LARS (T)          &\textbf{0.8496}	&\textbf{0.8789}	&\textbf{0.8545}&\textbf{0.8231}\\
&LARS (N)          &0.8102	&0.8549	&0.8285&0.7889\\
&LARS (T+N)        &0.8483	&0.8758	&0.8454&0.8159\\
\bottomrule
\end{tabular}
\caption{OOD data experiments on TriviaQA dataset with AUROC score. LARS models are trained with TriviaQA~(T) and/or NaturalQA (N).}
\vskip -0.1in
\label{tab:ood_trivia}
\end{table*}

\begin{figure*}[!htbp]
\begin{center}
\includegraphics[width=\textwidth]{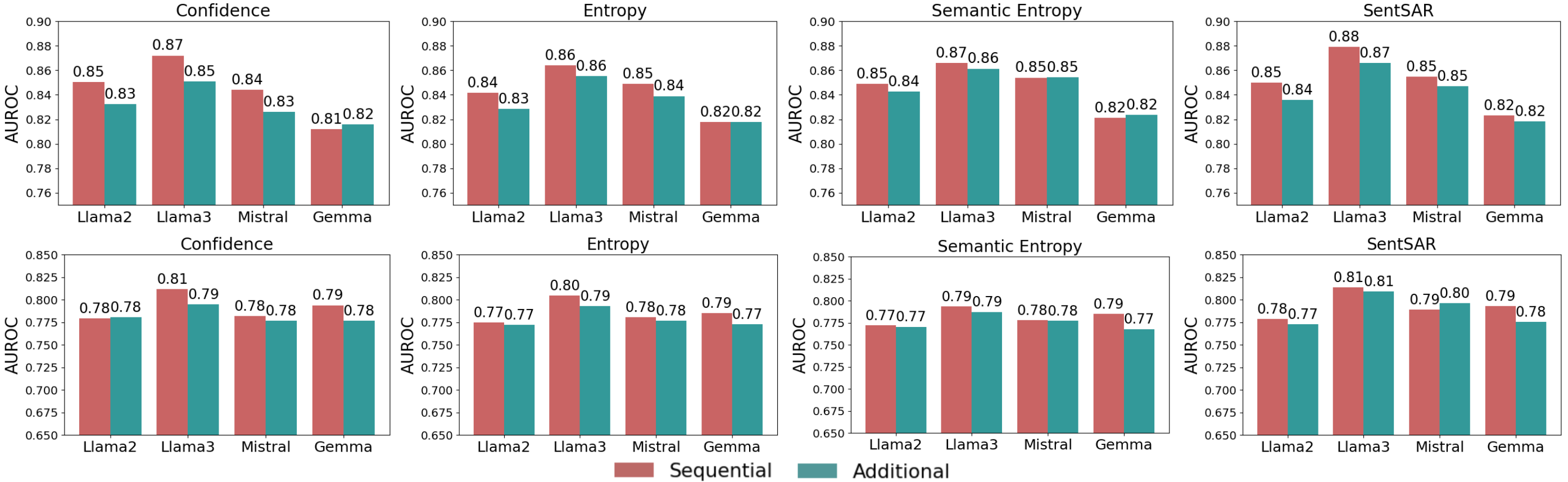}
\vskip -0.1in
\caption{Comparison of different probability association methods for LARS on TriviaQA (top) and NaturalQA (bottom).}
\label{fig:v1_v2_big}
\end{center}
\vskip -0.25in
\end{figure*}

\subsection{Extension of the Main Results}\label{app:main_ext}
Extended version of the main results are presented in Table \ref{tab:main_results_big} and \ref{tab:gsm8k}. 

In GSM8K, we observe a decrease in consistency-based methods, which is due to the sentence similarity step they include. When numeric values remain more important, sentence similarity models may perform worse, thus leading to lower performance of UE methods compared to general-knowledge datasets. Moreover, we see a common behavior that SE and sentSAR mostly improves performance compared to confidence and entropy on most of the scoring functions for TriviaQA, NaturalQA and WebQA. This increase is expected due to their mechanism of checking the consistency of the outputs. However, when the performance of sentence similarity measuring approaches are not stable–as we see in GSM8K–the positive effect of SE and SentSAR remains very low. A similar discussion can be made for Eccentricity and Degree Matrix approaches since they use the same sentence similarity model as SE.

\subsection{Probability Association Strategies}\label{app:prob_asc}
The extended comparison between sequential and additive probability association strategies are presented in Figure \ref{fig:v1_v2_big}.

\subsection{OOD Data Experiments -  LARS}\label{app:ood_data}
Extended OOD data results for WebQA and GSM8K are presented in Tables \ref{tab:webqa_big} and \ref{tab:gsm8k}.

Table \ref{tab:ood_natural} details OOD data experiments on NaturalQA, and Table \ref{tab:ood_trivia} covers OOD data experiments on TriviaQA. Training LARS with data from different distributions results in a performance drop. However, when we integrate the original calibration data with OOD data, LARS achieves better results in NaturalQA experiments. This suggests that increasing the dataset size, even with data from other distributions, might enhance the performance of LARS depending on the dataset.

\subsection{LARS without Labeled Data}\label{sec:teacher}

In this section, we explore the performance of LARS in the absence of labeled data. For this, for each question in the calibration dataset, we first use Llama3-8b to generate answers. To assess the correctness of these answers, we employ a teacher LLM (either Llama3-70b or Llama3-8b) and prompt it to evaluate the correctness of the generated answers. This method produces noisy labels, some of which are incorrect.

Despite these noisy labels, training LARS with them yields a good performance, surpassing both other baselines and the self-evaluation of the LLM (see Table~\ref{tab:teacher}). This finding is promising and suggests that the pre-trained nature of the RoBERTa model, which already possesses some understanding of textual inputs, enables it to understand key features from the noisy and partial feedback provided by the teacher LLM. This capability contributes to getting a better scoring function than asking the LLM itself. Such effectiveness of pre-trained models in handling noisy labels supports previous research \cite{kim2021envbert}, underscoring the potential of LARS for further investigation in such environments.

\begin{table}[!htbp]
\centering
\vskip -0.1in
\fontsize{8.5}{9}\selectfont
\begin{tabular}{c| cc }
\toprule
 & \multicolumn{2}{c}{Teacher Model} \\
\cmidrule{2-3}
UE Method & Llama3-70b & Llama3-8b  \\
\midrule[\heavyrulewidth]
Ask LLM  &0.746 & 0.635 \\
LARS (No Labeled Data)       & \textbf{0.837} & \textbf{0.809}\\
\bottomrule
\end{tabular}
\caption{Results for LARS trained without labeled data on TriviaQA. The Confidence method is used for UE. }
\vskip -0.25in
\label{tab:teacher}
\end{table}

\subsection{Effect of the Model Family Choice}\label{sec:transformer}

The reasoning behind our model choice for LARS is thoroughly explained in Section \ref{method}. To further validate our decision to use a transformer-based architecture, we trained a supervised MLP model that transforms the input text, output text, and probabilities into a fixed vector. Specifically, we used a sentence encoder to encode the text into a fixed vector and appended the corresponding probabilities, which served as input for the MLP.

The results, presented in Table \ref{tab:mlp}, clearly demonstrate that LARS consistently outperforms the MLP by substantial margins. These findings indicate that a naive input strategy, such as the one used for the MLP, fails to capture the complexities of the problem, whereas a more sophisticated model family, like the one employed by LARS, is necessary to achieve optimal performance in uncertainty estimation. Moreover, the input strategy used in the MLP performs worse than directly feeding the probability vectors into the model. This could be due to the fact that adding a fixed representation of the text meaning increases the input dimensionality without significantly benefiting UE. As a result, this approach may reduce the model's ability to generalize effectively.

\begin{table}[!htbp]
\vskip -0.1in
\centering
\fontsize{8.5}{8}\selectfont
\begin{tabular}{c|l| cc }
\toprule
{UE Method}& {Scoring Function} & {AUROC} & PRR\\
\midrule[\heavyrulewidth]
\multirow{2}{*}{\textbf{Confidence}}
&MLP      &0.666	&0.398  \\
&LARS            &\textbf{0.851} &\textbf{ 0.760}\\
\midrule[\heavyrulewidth]
\multirow{2}{*}{\textbf{Entropy}}
&MLP       &0.718 &0.509 	  \\
&LARS            &\textbf{0.842} &\textbf{0.748} \\
\midrule[\heavyrulewidth]
\multirow{2}{*}{\textbf{SE}}
&MLP       &0.787 &0.634   \\
&LARS            &\textbf{0.849} &\textbf{0.745} 	\\
\midrule[\heavyrulewidth]
\multirow{2}{*}{\textbf{SentSAR}}
&MLP        &0.744   &0.571 \\
&LARS            &\textbf{0.850} &\textbf{0.763} 	\\
\bottomrule
\end{tabular}
\vskip -0.1in
\caption{ Comparison of LARS and MLP with the same modalities on Llama2-7b model on the TriviaQA.}

\vskip -0.2in
\label{tab:mlp}
\end{table}

\subsection{Choice of Encoder-only Transformer}\label{app:lars_model}
\begin{figure*}[!htbp]
\begin{center}
\includegraphics[width=0.7\textwidth]{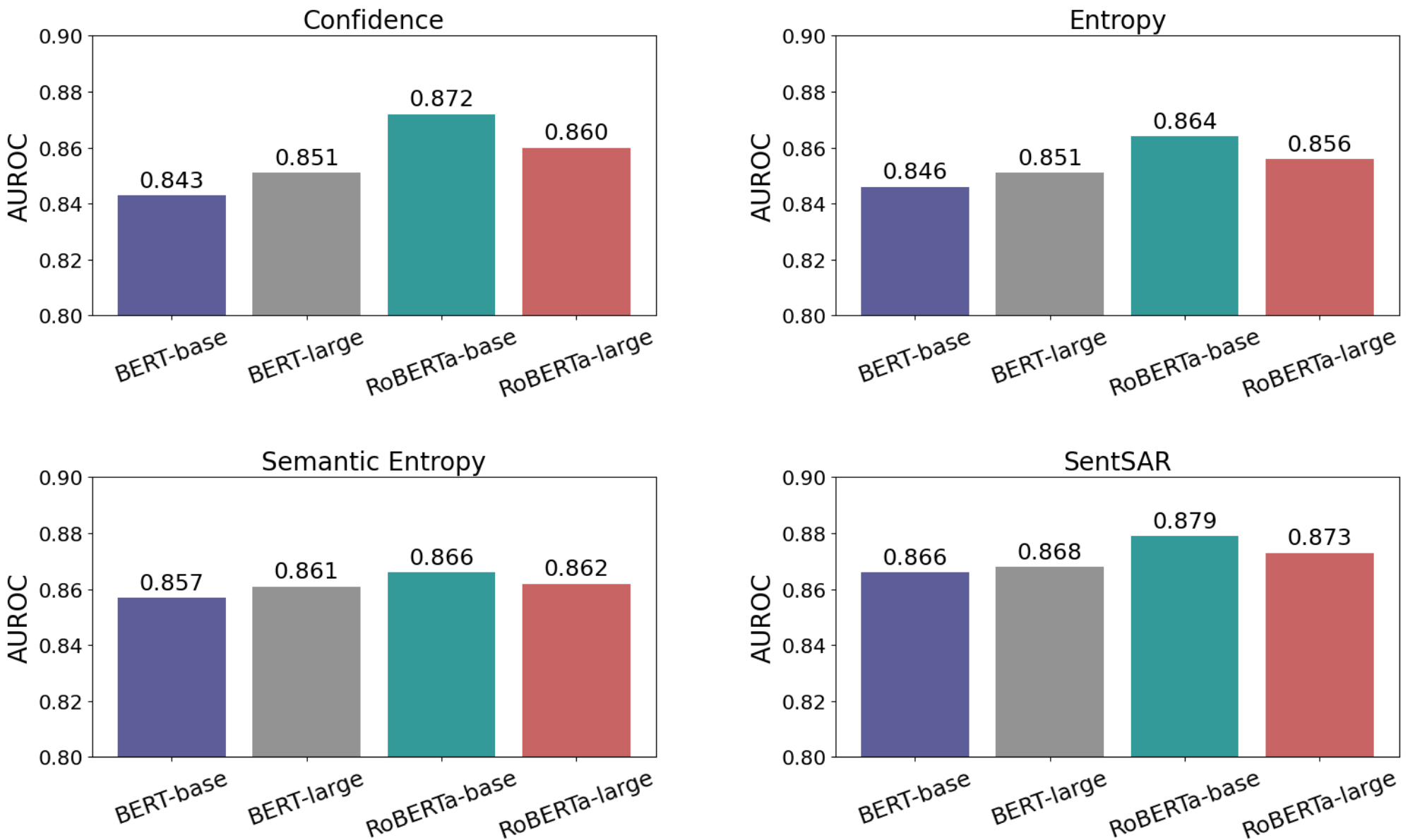}
\caption{AUROC scores for different choice of LARS models on TriviaQA and LLama3-8b.}
\label{fig:lars_models}
\end{center}
\end{figure*}

To evaluate the effect of LARS model selection on both architecture and model size, we trained four LARS models using TriviaQA-LLama3-8b calibration data. The LARS models utilized are: bert-base-uncased, bert-large-uncased \cite{devlin2019bert}, roberta-base, and roberta-large \cite{liu2019roberta}. The sizes of each model are 110M, 336M, 125M, and 355M, respectively. The results are presented in Figure \ref{fig:lars_models}. When comparing BERT and RoBERTa models of similar sizes, it is evident that RoBERTa consistently outperforms BERT. As model size increases, BERT's performance improves, whereas RoBERTa exhibits the opposite behavior. Notably, a detailed hyperparameter search was not performed for RoBERTa-large. If conducted, this might allow RoBERTa-large to surpass RoBERTa-base; however, considering the inference costs, RoBERTa-base is used as the default LARS model.

\subsection{Effect of Number of Probability Tokens}

\begin{figure*}[!htbp]
\begin{center}
\includegraphics[width=\textwidth]{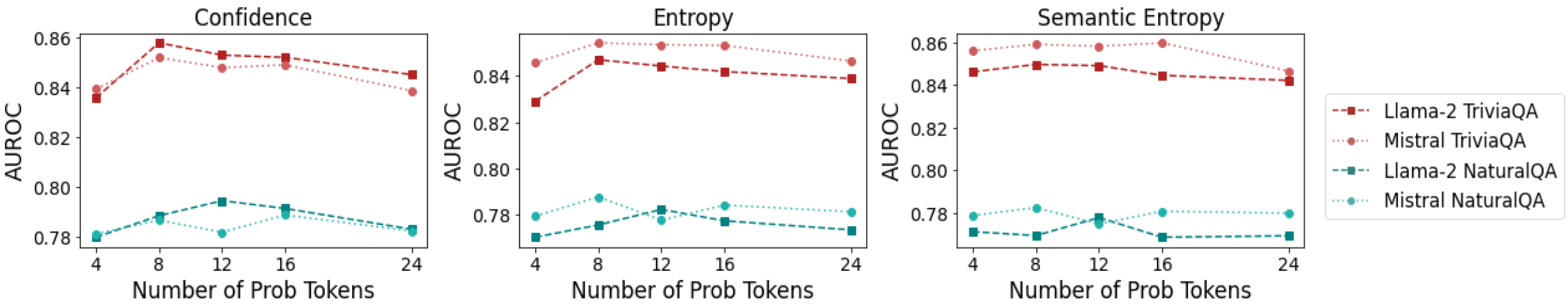}
\caption{AUROC scores for varying number of probability tokens for LARS on 2 models and 2 datasets.}
\label{fig:num_of_bins}
\end{center}
\end{figure*}

Figure~\ref{fig:num_of_bins} shows the impact of varying the number of probability tokens, $k$ during LARS training.
Probabilities are divided into $k$ quantiles, each represented by a unique few-hot vector, as described in Section \ref{method}.
The choice of $k$ directly influences the bias-variance trade-off of the model. With a high number of probability tokens, the model may overfit, reflecting minor fluctuations in probability within the inputs. Conversely, a small number of tokens might hinder the model’s ability to distinguish between significantly different probabilities, as they are represented by identical tokens. Our results indicate that using 8 quantiles for the probability vectors generally yields the best generalization.

\subsection{LARS on Different Languages}\label{app:lang}
Extended results on different languages from Section \ref{turkish} are presented in Table \ref{tab:lang_big}. In all languages, LARS consistently outperforms baseline UE methods and scoring functions.

\subsection{Impact of Question as LARS Input}\label{app:no_question}
In this section, we evaluate the impact of including the question as input to the LARS model. The results, shown in Table \ref{tab:no_question}, indicate a consistent performance drop when the question is omitted. This outcome is expected, as the question context plays a crucial role in determining whether a generated response is nonsensical or off-topic. However, we argue that the performance drop is not substantial. Thus, if computational efficiency is a priority, LARS can still be effectively used without the question context.

\begin{table}[!htbp]
\centering
\fontsize{8.5}{7.5}\selectfont
\begin{tabular}{c|l| ccc }
\toprule
{UE Method}& {Scoring Function} & {AUROC} & PRR\\
\midrule[\heavyrulewidth]
\multirow{2}{*}{\textbf{Confidence}}
&No question     &0.832   &0.720 \\
&LARS            &\textbf{0.851} & \textbf{0.760}\\
\midrule[\heavyrulewidth]
\multirow{2}{*}{\textbf{Entropy}}
&No question     &0.828   &0.716 \\
&LARS            &\textbf{0.842} &\textbf{0.748} \\
\midrule[\heavyrulewidth]
\multirow{2}{*}{\textbf{SE}}
&No question     &0.836   &0.736 \\
&LARS            &\textbf{0.849} &\textbf{0.745} 	\\
\midrule[\heavyrulewidth]
\multirow{2}{*}{\textbf{SentSAR}}
&No question     &0.842   &0.740 \\
&LARS            &\textbf{0.850} &\textbf{0.763} 	\\
\bottomrule
\end{tabular}
\caption{ LARS with and without question in the input.}
\label{tab:no_question}
\end{table}

\begin{table*}[!htbp]
\centering
\fontsize{9.2}{9}\selectfont
\begin{tabular}{c|l| cc |cc|cc|cc }
\toprule
  UE Method & Scoring & \multicolumn{2}{c}{\textbf{English}} & \multicolumn{2}{c}{\textbf{Turkish}} & \multicolumn{2}{c}{\textbf{German}}& \multicolumn{2}{c}{\textbf{Spanish}} \\
   & Function & AUROC & PRR & AUROC & PRR & AUROC & PRR & AUROC & PRR \\
\midrule[\heavyrulewidth]

\textbf{Lex. Sim.}   & -         &0.683 &0.483 &0.652 &0.361 &0.660 &0.410 &0.640 &0.369  \\
\textbf{\# Sem. Gr.} & -         &0.819 &0.671 &0.683 &0.343 &0.774 &0.571 &0.787 &0.605  \\
\textbf{Eccentricity} & -        &0.853 &0756. &0.732 &0.482 &0.805 &0.668 &0.813 &0.689  \\
\textbf{Degree Matrix} & -       &0.851 &0.746 &0.748 &0.583 &0.802 &0.675 &0.814 &0.692  \\

\midrule[\heavyrulewidth]
\multirow{4}{*}{\textbf{Confidence}}
&LNS        &0.748 &0.600 &0.714 &0.500 &0.727 &0.544 &0.704 &0.504  \\
&MARS       &0.799 &0.676 &0.720 &0.503 &0.747 &0.582 &0.728 &0.550  \\
&TokenSAR   &0.792 &0.674 &0.747 &0.568 &0.779 &0.645 &0.761 &0.609  \\
&LARS       &\textbf{0.872} &\textbf{0.817} &\textbf{0.831} &\textbf{0.703} &\textbf{0.843} &\textbf{0.754} &\textbf{0.852} &\textbf{0.764}  \\
\midrule[\heavyrulewidth]
\multirow{4}{*}{\textbf{Entropy}}
&LNS        &0.747 &0.594 &0.692 &0.450 &0.710 &0.523 &0.701 &0.491  \\
&MARS       &0.801 &0.672 &0.695 &0.457 &0.728 &0.555 &0.723 &0.533  \\
&TokenSAR   &0.793 &0.676 &0.720 &0.515 &0.758 &0.614 &0.750 &0.590  \\
&LARS       &\textbf{0.864} &\textbf{0.804} &\textbf{0.814} &\textbf{0.680} &\textbf{0.827} &\textbf{0.737} &\textbf{0.835} &\textbf{0.742}  \\
\midrule[\heavyrulewidth]
\multirow{4}{*}{\textbf{SE}}
&LNS        &0.835 &0.733 &0.734 &0.554 &0.791 &0.663 &0.797 &0.667  \\
&MARS       &0.835 &0.742 &0.734 &0.551 &0.789 &0.663 &0.795 &0.666  \\
&TokenSAR   &0.839 &0.747 &0.739 &0.568 &0.796 &0.676 &0.800 &0.677  \\
&LARS       &\textbf{0.866} &\textbf{0.811} &\textbf{0.799} &\textbf{0.668} &\textbf{0.824} &\textbf{0.735} &\textbf{0.831} &\textbf{0.742}  \\
\midrule[\heavyrulewidth]
\multirow{4}{*}{\textbf{SentSAR}}
&LNS        &0.825 &0.723 &0.728 &0.530 &0.765 &0.629 &0.765 &0.617  \\
&MARS       &0.838 &0.746 &0.731 &0.531 &0.775 &0.641 &0.775 &0.631  \\
&TokenSAR   &0.840 &0.750 &0.752 &0.577 &0.793 &0.673 &0.790 &0.660  \\
&LARS       &\textbf{0.879} &\textbf{0.823} &\textbf{0.828} &\textbf{0.705} &\textbf{0.841} &\textbf{0.757} &\textbf{0.848} &\textbf{0.763}  \\
\bottomrule
\end{tabular}
\caption{AUROC and PRR performance of different UE methods on Llama3-8B for the TriviaQA dataset in different languages.}
\vskip -0.2in
\label{tab:lang_big}
\end{table*}

\subsection{OOD Model Experiments -  LARS}\label{app:ood_model}
In this section, we present extensive OOD model experiments for LARS. The results are detailed in Table \ref{tab:ood_model}, with interpretations similar to those in Table \ref{tab:ood_model_small}. Training LARS on outputs from different LLMs results in an expected performance drop. Nonetheless, LARS continues to outperform other scoring functions, demonstrating its robustness and potential.

In this experiment, for each LLM we use, we train a LARS model using all of the TriviaQA and NaturalQA samples we created for training.

\begin{table*}[!htbp]
\centering
\fontsize{9.3}{9}\selectfont
\begin{tabular}{c |c|l| ccccc }
\toprule
 Dataset&UE Method & Calibration Model & \textbf{Llama2-7b} & \textbf{Llama3-8b} & \textbf{Mistral-7b}& \textbf{Gemma-7b} & \textbf{Llama2-13b} \\
\midrule[\heavyrulewidth]

\multirow{24}{*}{\rotatebox{90}{\textbf{TriviaQA}}}
&\multirow{6}{*}{\textbf{Confidence}}
&Best Baseline Score       &0.7510	&0.7994	&0.7468	&0.6883 &0.7278  \\
&&Llama2-7b     	&\textbf{0.8577}	&0.8355	&0.8309	&0.7993 &0.8432 \\
&&Llama3-8b     	&0.8519	&\textbf{0.8737}	&0.8499	&0.7830 &0.8146 \\
&&Mistral-7b        &0.8352	&0.8327	&\textbf{0.8518}	&0.7872 &0.8315 \\
&&Gemma-7b      	&0.8169	&0.8172	&0.8097	&\textbf{0.8311}	&0.8054 \\
&&Llama2-13b      	&0.8376	&0.8526	&0.8327	&0.7829 &\textbf{0.8510} \\
\cmidrule{2-8}
&\multirow{6}{*}{\textbf{Entropy}}
&Best Baseline Score       &0.7356	&0.8012	&0.7634	&0.6941 &0.7332  \\
&&Llama2-7b     	&0.8469	&0.8298	&0.8271	&0.8060	&0.8473 \\
&&Llama3-8b     	&\textbf{0.8520}	&\textbf{0.8730}	&0.8501	&0.7955	&0.8313 \\
&&Mistral-7b        &0.8410	&0.8407	&\textbf{0.8542}	&0.7974	&0.8380 \\
&&Gemma-7b      	&0.8145	&0.8181	&0.8200	&\textbf{0.8279}	&0.8115 \\
&&Llama2-13b      	&0.8335	&0.8525	&0.8435	&0.7916 &\textbf{0.8553} \\
\cmidrule{2-8}
&\multirow{6}{*}{\textbf{SE}}
&Best Baseline Score       &0.7973	&0.8388	&0.8132	&0.7528 &0.8062 \\
&&Llama2-7b     	&0.8497	&0.8358	&0.8402	&0.8100	&0.8603 \\
&&Llama3-8b     	&\textbf{0.8625}	&\textbf{0.8719}	&\textbf{0.8623}	&0.8008	&0.8518 \\
&&Mistral-7b        &0.8496	&0.8490	&0.8591	&0.8056	&0.8579 \\
&&Gemma-7b      	&0.8334	&0.8425	&0.8372	&\textbf{0.8289}	&0.8386 \\
&&Llama2-13b      	&0.8447	&0.8660	&0.8554	&0.8049 &\textbf{0.8671} \\
\cmidrule{2-8}
&\multirow{6}{*}{\textbf{SentSAR}}
&Best Baseline Score       &0.7940   &0.8402   &0.8050  &0.7411 &0.7910 \\
&&Llama2-7b     	&0.8572	&0.8409	&0.8413	&0.7900	&0.8584 \\
&&Llama3-8b     	&\textbf{0.8663}	&\textbf{0.8838}	&\textbf{0.8634}	&0.8017	&0.8453 \\
&&Mistral-7b        &0.8514	&0.8474	&0.8596	&0.8055	&0.8513 \\
&&Gemma-7b      	&0.8309	&0.8363	&0.8327	&\textbf{0.8338}	&0.8272 \\
&&Llama2-13b      	&0.8456	&0.8687	&0.8506	&0.7980 &\textbf{0.8616} \\
\midrule[\heavyrulewidth]
\multirow{24}{*}{\rotatebox{90}{\textbf{NaturalQA}}}

&\multirow{6}{*}{\textbf{Confidence}}
&Best Baseline Score       &0.7032	&0.7136	&0.6823	&0.6433  &0.6774 \\
&&Llama2-7b     	&\textbf{0.7886}	&0.7546	&0.7512	&0.7288	&0.7613 \\
&&Llama3-8b     	&0.7732	&\textbf{0.8113}	&0.7679	&0.7265	&0.7517 \\
&&Mistral-7b        &0.7538	&0.7543	&\textbf{0.7868}	&0.7195	&0.7507 \\
&&Gemma-7b      	&0.7522	&0.7397	&0.7493	&\textbf{0.8033}	&0.7295 \\
&&Llama2-13b      	&0.7727	&0.7695	&0.7571	&0.7308 &\textbf{0.7812} \\
\cmidrule{2-8}
&\multirow{6}{*}{\textbf{Entropy}}
&Best Baseline Score       &0.6831	&0.7144	&0.6944	&0.6439 &0.6859  \\
&&Llama2-7b     	&\textbf{0.7756}	&0.7582	&0.7550	&0.7367	&0.7641 \\
&&Llama3-8b     	&0.7734	&\textbf{0.8103}	&0.7767	&0.7374	&0.7544 \\
&&Mistral-7b        &0.7569	&0.7642	&\textbf{0.7877}	&0.7305	&0.7607 \\
&&Gemma-7b      	&0.7481	&0.7463	&0.7506	&\textbf{0.7939}	&0.7351 \\
&&Llama2-13b      	&0.7671	&0.7752	&0.7569	&0.7363 &\textbf{0.7783} \\
\cmidrule{2-8}
&\multirow{6}{*}{\textbf{SE}}
&Best Baseline Score       &0.7210	&0.7591	&0.7272	&0.7049 &0.7361 \\
&&Llama2-7b     	&0.7695	&0.7590	&0.7574	&0.7521	&0.7716 \\
&&Llama3-8b     	&0.7767	&\textbf{0.8038}	&0.7820	&0.7513	&0.7661 \\
&&Mistral-7b        &0.7627	&0.7681	&\textbf{0.7826}	&0.7484	&0.7680 \\
&&Gemma-7b      	&0.7517	&0.7602	&0.7561	&\textbf{0.7916}	&0.7511 \\
&&Llama2-13b      	&\textbf{0.8049}	&0.7837	&0.7672	&0.7548 &\textbf{0.7843} \\
\cmidrule{2-8}
&\multirow{6}{*}{\textbf{SentSAR}}
&Best Baseline Score      &0.7177 &0.7563 &0.7268 &0.6891 &0.7319  \\
&&Llama2-7b     	&0.7835	&0.7639	&0.7595	&0.7423	&0.7738 \\
&&Llama3-8b     	&0.7816	&\textbf{0.8154}	&0.7838	&0.7417	&0.7655 \\
&&Mistral-7b        &0.7669	&0.7705	&\textbf{0.7940}	&0.7360	&0.7682 \\
&&Gemma-7b      	&0.7572	&0.7567	&0.7616	&\textbf{0.7978}	&0.7486 \\
&&Llama2-13b      	&\textbf{0.7980}	&0.7838	&0.7654	&0.7415 &\textbf{0.7853} \\
\bottomrule
\end{tabular}
\caption{OOD model experiments on TriviaQA and NaturalQA datasets with AUROC scores.}
\vskip -0.1in
\label{tab:ood_model}
\end{table*}

\begin{table*}[!htbp]
\vskip 0.3in

\centering
\fontsize{4.2}{3.4}\selectfont
\resizebox{\textwidth}{!}{%
\begin{tabular}{c|p{5cm}|c}
\toprule
&\multicolumn{1}{c|}{ \textbf{Question}} &\multicolumn{1}{c}{ \textbf{Ground Truth}}  \\
\midrule[\heavyrulewidth]

\multirow{8}{*}{\rotatebox{90}{\textbf{TriviaQA}}}
&\multicolumn{1}{m{4cm}|}{David Lloyd George was British Prime Minister during the reign of which monarch?}&
King George V\\
\cmidrule{2-3}
&\multicolumn{1}{m{4cm}|}{How many symphonies did Jean Sibelius compose?}&
Seven\\
\cmidrule{2-3}
&\multicolumn{1}{m{4cm}|}{The capital of Brazil was moved from Rio de Janeiro to the purpose-built capital city of Brasilia in what year?}&
1960\\

\midrule[\heavyrulewidth]
\multirow{3}{*}{\rotatebox{90}{\textbf{NaturalQA}}}
&\multicolumn{1}{m{4cm}|}{when was the last time anyone was on the moon}&December 1972\\
\cmidrule{2-3}
&\multicolumn{1}{m{4cm}|}{who wrote he ain't heavy he's my brother lyrics}&
Bobby Scott, Bob Russell\\
\cmidrule{2-3}
&\multicolumn{1}{m{4cm}|}{how many seasons of the bastard executioner are there}&one\\

\midrule[\heavyrulewidth]
\multirow{6}{*}{\rotatebox{90}{\textbf{WebQA}}}
&\multicolumn{1}{m{4cm}|}{what is the name of justin bieber brother?}&Jazmyn Bieber\\
\cmidrule{2-3}
&\multicolumn{1}{m{4cm}|}{what character did natalie portman play in star wars?}&
Padmé Amidala\\
\cmidrule{2-3}
&\multicolumn{1}{m{4cm}|}{what character did john noble play in lord of the rings?}&
Denethor II\\

\midrule[\heavyrulewidth]
\multirow{13}{*}{\rotatebox{90}{\textbf{GSM8K}}}
&\multicolumn{1}{m{4cm}|}{Natalia sold clips to 48 of her friends in April, and then she sold half as many clips in May. How many clips did Natalia sell altogether in April and May?}&72\\
\cmidrule{2-3}
&\multicolumn{1}{m{4cm}|}{Julie is reading a 120-page book. Yesterday, she was able to read 12 pages and today, she read twice as many pages as yesterday. If she wants to read half of the remaining pages tomorrow, how many pages should she read?}&
42\\
\cmidrule{2-3}
&\multicolumn{1}{m{4cm}|}{Mr. Sam shared a certain amount of money between his two sons, Ken and Tony. If Ken got \$1750, and Tony got twice as much as Ken, how much was the money shared?}&
5250\\

\bottomrule
\end{tabular}}
\caption{Data samples from the datasets we use to evaluate UE methods: TriviaQA, NaturalQA, WebQA, and GSM8K. }
\label{tab:data_samples}
\end{table*}

\section{Experimental Details} \label{exp_details}

\noindent\textbf{Datasets.} To train the LARS model, for each TriviaQA and NaturalQA training split, we randomly select $\sim$13k samples resulting in $\sim$60k sampled unique QA pairs. We use all of the train split of GSM8K containing $\sim$8k samples. To evaluate the UE methods we use 4 datasets: $\sim$9k samples from the TriviaQA validation split, the validation set of NaturalQA consisting of $\sim$3500 samples, $\sim$6k samples coming from the train and validation sets of WebQA combined, and complete test split of GSM8K containing $\sim$1k samples.

\bigskip

\noindent \textbf{Example Samples from Datasets.}
We provide samples from the datasets we use for the evaluation of UE methods in Table \ref{tab:data_samples}.

\bigskip

\noindent \textbf{Generation Configurations.} We utilize Huggingface library and its built-in $\texttt{generate()}$ function to obtain answers. We use $\texttt{num\_beams=1}$. For the most likely responses we set $\texttt{do\_sample=False}$ while for the set of sampled generations, it is True. We set the default LLMs' eos token as end of sentence token to stop the generation.

\bigskip

\noindent \textbf{GPT Performance in Evaluation of Deneration Correctness.}
Following prior works \cite{lin2023generating, tokensar, bakman2024mars}, we employed GPT-3.5-turbo to assign correctness labels to model-generated answers based on the provided ground truth and question. To assess the effectiveness of GPT-3.5-turbo in this task, we conducted a human evaluation.
A human evaluator independently assessed the answers against the ground truth and the question without access to GPT-generated labels. The accuracy of GPT-3.5-turbo's correctness labels was then calculated by comparing them to the human evaluations. It obtained an accuracy of 96\%, highlighting the high performance of GPT-3.5-turbo in this task.

\bigskip

\noindent \textbf{Computational Cost.} 
We use 40 GB Nvidia A-100 GPUs for all the experiments. The total GPU-hours for training a LARS model with a calibration dataset generated from $\sim$13k questions is approximately 4. Labeling of the calibration data for one dataset and one model takes approximately 30 GPU-hours. Getting all the results in Tables~\ref{tab:main_results_big} and \ref{tab:gsm8k} compromises $\sim$300 GPU-hours excluding LARS training. All presented results are obtained with a single run. 

\bigskip

\noindent \textbf{Prompts.}
The prompts for the LLM models to generate answers to questions are given below.

For LLama family:
\begin{verbatim}
System:You are a helpful, respectful 
and honest assistant. Give precise, 
short, one sentence answers to given 
questions. Do not use emojis. 
User:{question}
\end{verbatim}

For Mistral-7b:
\begin{verbatim}
User: Give precise, short, one 
sentence answers to given 
questions. {question}
\end{verbatim}

For Gemma-7b:
\begin{verbatim}
User: You are a helpful, respectful 
and honest assistant. Give precise, 
short, one sentence answers to 
given questions. Question:{question}
\end{verbatim}

\bigskip

The prompt used for GPT-3.5-turbo to obtain labels:
\begin{verbatim}
You will behave as a question answer 
evaluator. I will give you a question, 
the ground truth of the question, and 
a generated answer by a language model.
You will output "correct" if the 
generated answer is correct regarding 
question and ground truth. 
Otherwise, output "false".
Question: {question}, 
Ground Truth: {gt_answer},
Generated Answer: {generation}
\end{verbatim}

\bigskip

The prompt for the teacher models explained in Section \ref{sec:teacher} is as follows:
\begin{verbatim}
System: You are a helpful, respectful 
and honest question-answer evaluator. 
You will be given a question and a 
possible answer. Evaluate the 
possible answer as true or false 
considering the question. Output 
"true" if the answer is correct. 
Otherwise, output "false". Do not 
make any explanation.
User: Question:{question}
Possible answer:{answer}
\end{verbatim}

\bigskip

The prompts for the LLM models to self-check their answers for $p$(True) evaluation is provided below.
For Llama family:
\begin{verbatim}
System: You are a helpful, respectful 
and honest question-answer evaluator. 
You will be given a question, some 
brainstormed ideas and a possible 
answer. Evaluate the possible answer 
as True or False considering the 
question and brainstormed ideas. 
Output only True or False.
User: Question:{few_shot_q1}
Here are some ideas that were 
brainstormed:{few_shot_samples1}
Possible answer:{few_shot_ans1}
The possible answer is: 
Assistant: True
User: Question:{few_shot_q2}
Here are some ideas that were 
brainstormed:{few_shot_samples2}
Possible answer:{few_shot_ans2}
The possible answer is: 
Assistant: False
User: Question:{question}
Here are some ideas that were 
brainstormed:{sampled_generation}
Possible answer:{most_likelt_gen}
The possible answer is: 
\end{verbatim}


For Mistral-7b and Gemma-7b:
\begin{verbatim}
User: You are a helpful, respectful 
and honest question-answer evaluator. 
You will be given a question, some 
brainstormed ideas and a possible 
answer. Evaluate the possible answer 
as True or False considering the 
question and brainstormed ideas. 
Output only True or False.
Question:{few_shot_q1}
Here are some ideas that were 
brainstormed:{few_shot_samples1}
Possible answer:{few_shot_ans1}
The possible answer is: 
Assistant: True
User: Question:{few_shot_q2}
Here are some ideas that were 
brainstormed:{few_shot_samples2}
Possible answer:{few_shot_ans2}
The possible answer is: 
Assistant: False
User: Question:{question}
Here are some ideas that were 
brainstormed:{sampled_generation}
Possible answer:{most_likelt_gen}
The possible answer is: 
\end{verbatim}

\end{document}